\documentclass[preprint,12pt]{elsarticle}

\usepackage{amssymb}
\usepackage{amsmath}

\journal{Information Sciences}

\myfooter[L]{\parbox[t]{\textwidth}{\raggedright\upshape\urlstyle{same}Accepted manuscript. Published in Information Sciences 766 (2027) 124394,
\url{https://doi.org/10.1016/j.ins.2026.124394}.\\
\copyright~2026. This manuscript version is made available under the CC BY-NC-ND 4.0 license
\url{https://creativecommons.org/licenses/by-nc-nd/4.0/}.}}

\usepackage{amsfonts, amsthm} 
\usepackage{ifthen,tikz} 
\usepackage{lineno} 
\usepackage{pgfplots} 
\pgfplotsset{compat=1.17}
\usepackage{xcolor}
\usepackage{colortbl}

\newif\ifshowrevisions
\showrevisionsfalse
\ifshowrevisions
  \usepackage{soul}
  \sethlcolor{yellow}
  \newcommand{\rev}[1]{\begingroup\sethlcolor{yellow}\color{black}\hl{#1}\endgroup}
  
\else
  \newcommand{\rev}[1]{#1}
  
\fi

\usepackage{hyperref}
\hypersetup{colorlinks=true, linkcolor=blue!60!black, citecolor=blue!60!black, urlcolor=blue!60!black}
\ifshowrevisions
  \usepackage{pdfcomment}
  \newcommand{\rcmark}[1]{\pdfcomment[icon=Comment]{#1}}
\else
  \newcommand{\rcmark}[1]{}
\fi
\usepackage{ulem}

\pgfplotsset{
  ml/.style={thick,solid},
  nn/.style={thick,dashed},
grpOne/.style={blue,mark=o},
  grpEight/.style={red,mark=square},
  grpSixteen/.style={green!60!black,mark=triangle},
  mymarks/.style={mark size=2.2pt, mark options={solid,fill=white}}, }

\usetikzlibrary{positioning, arrows.meta, shapes.geometric}

\newtheorem{theorem}{Theorem}
\newtheorem{lemma}{Lemma} 
\newtheorem{corollary}{Corollary}

\begin{document}

\begin{frontmatter}

\title{Neural Classifiers as Error-Correcting Codes}

\author[1]{Sergey Salishev\corref{cor1}}
\ead{s.salischev@spbu.ru}
\author[1]{Anton Makarov}
\ead{a.a.makarov@spbu.ru}
\author[1]{Oleg Granichin}
\ead{o.granichin@spbu.ru}
\author[2]{Renata Avros}
\ead{ravros@braude.ac.il}
\author[2]{Zeev Volkovich}
\ead{vlvolkov@braude.ac.il}
\cortext[cor1]{Corresponding author}

\affiliation[1]{organization={St. Petersburg State University},
city={Saint Petersburg}, postcode={199034},
country={Russia}}
\affiliation[2]{organization={Department of Software Engineering, Braude College},
postcode={21982},
city={Karmiel},
country={Israel}}

\begin{abstract}
A unified framework is proposed that characterizes supervised deep classification as the decoding of error-correcting codes 
in the Shannon sense. Cross-entropy optimization is shown to align softmax classifiers with maximum-likelihood decoding, where 
logits map to Hamming distances to class-specific codewords. This correspondence explains the emergence of standard divergence 
objectives (e.~g., Kullback-Leibler, Jeffreys, Bregman) and provides a geometric basis for the stability of decision boundaries. 
Furthermore, the geometries of Neural Collapse and LeJEPA are identified as optimal constellations for decoding under isotropic noise. 
Unlike classical Error-Correcting Output Codes, deep networks are shown to learn effective codebooks in the feature space implicitly. 
This insight is leveraged to derive novel criteria for lossless quantization, improved knowledge distillation, and architectural 
regularization based on code redundancy. Finally, this perspective is extended to self-supervised encoders, offering a cohesive 
explanation for the geometry of latent representations in modern neural models.
\end{abstract}


\end{frontmatter}

\section{Introduction} \label{sec:intro}
Convolutional Neural Networks (CNNs) are ubiquitous, de facto standards for solving complex classification problems across many areas, such as computer vision, speech, and text processing. A mainstream architecture comprises an arbitrary number of nonlinear layers that aim to provide a suitable data representation at sequentially increasing levels of abstraction. Classification is performed using fully connected layers, capped by a softmax layer. The system is optimized using the \rev{cross-entropy loss function}~\cite{Goodfellow-et-al-2016}. Such a setup implicitly assumes a probabilistic interpretation of the outputs as \rev{posterior class probabilities}~\cite{Bridle-1990-Softmax} and is closely related to the Kullback-Leibler divergence. At the same time, the connection between the standard deep learning practice and the classical theory of information transmission and forward error-correcting (FEC) codes remains at the level of heuristic analogies.

Classical information theory asserts that optimal recognition procedures are fundamentally equivalent to decoders tasked with reconstructing signals distorted by a \rev{noisy communication channel}~\cite{MacKay-2003-ITILA}, a principle rigorously grounded in Shannon's theorems on channel capacity and the guaranteed existence of efficient \rev{error-correcting codes}~\cite{Cover-Thomas-IT-2006}. This theoretical correspondence naturally raises an important question for contemporary deep learning: to what extent can a neural classifier be interpreted as a practical decoder of an FEC code, with its intermediate internal states functioning as codewords degraded by computational noise and quantization? Clarifying the limits of this coding-theoretic analogy is essential, as affirming its validity could offer significant implications for the design, optimization, training dynamics, and quantization methods used in modern neural networks.

Interpreting neural classifiers as stochastic decision systems aligns with earlier work on randomized algorithms in automatic
control and data mining, where robustness and performance guarantees are derived from probabilistic modeling of uncertainty and
noise~\cite{GranichinVolkovich2015}.

A complementary channel-based perspective on quantization is given in \cite{salishev2025gdnsq}.
While it focuses on quantization viewed as transmission over a noisy binary channel, it does not address the decoding interpretation of supervised classification or the geometry of class representations, which are the focus of the present work.

A direct connection between multiclass classification and error-correcting codes has been studied in detail within the error-correcting output codes (ECOC) \rev{methodology}~\cite{Dietterich-Bakiri-1995-ECOC}, where each class is associated with a discrete binary code, and the multiclass classification problem is treated as training a set of binary classifiers on the bits of this code, followed by decoding by Hamming distance \rev{or, for margin classifiers, by loss-based decoding}~\cite{Allwein-2000-ReducingMulticlass}. This approach explicitly borrows ideas from coding theory and demonstrates that a well-chosen code matrix increases the classifier's robustness to errors in individual decision ``bits''.

However, the ECOC approach is essentially limited to the output representation: the codes are set at the last layer, and the deep network's internal structure is treated as a ``black box'' that produces features for these bit classifiers. In addition, the code matrix is chosen explicitly (or by simple heuristics) and is not interpreted as the result of random coding in the sense of Shannon's theorems. Finally, the loss functions used are usually selected for practical reasons (e.~g., cross-entropy, hinge loss), rather than explicitly derived from optimal decoding criteria.

Recent studies on the Neural Collapse phenomenon have shown that under the standard practice of training with cross-entropy and a softmax output, the last layers of deep networks possess a surprisingly simple geometry. \rev{Papyan et al.}~\cite{papyan2020neuralcollapse} \rev{and Han et al.}~\cite{Han-2021-NC-MSE} have shown that in the terminal phase of training, features within each class collapse to their means, and the class means form the vertices of a simplex. Moreover, the classifier vectors also align in the dual configuration, coinciding (up to scale) with these means. This structure directly resembles the geometry of optimal spherical codes and dense packings well known in coding theory.

Neural Collapse (NC) refers to a striking and highly structured set of behaviors consistently observed in deep neural networks during the final stages of training for classification problems. This phenomenon is particularly pronounced when networks are trained with the cross-entropy loss function and a softmax output layer. Formally introduced in~\cite{papyan2020neuralcollapse}, NC describes a series of geometric regularities that reliably emerge in both the learned feature representations (what the network extracts from the data) and the classifier parameters as the network approaches a state of zero training error.

But generally speaking, NC does not carry the analogy with FEC codes through to a formal equivalence between classification and decoding tasks under noise and the maximum likelihood (ML) probabilistic criteria. In particular, the question remains open as to how the standard cross-entropy loss and the Hamming distance between some ``hidden'' binary features are related, as well as how to interpret the softmax layer and its logits from the standpoint of optimal code decoding.

A concurrent research direction in self-supervised learning focuses on treating the representations learned by neural networks as encodings of the input data within a latent space that possesses an explicitly defined geometry. This geometric approach is implemented in Joint-Embedding Predictive Architectures (JEPA) and its recent derivative, LeJEPA~\cite{Balestriero-2025-LeJEPA}. JEPA learns by predicting target embeddings from partial observations rather than by reconstructing the input itself, and LeJEPA is the corresponding latent-embedding formulation with an explicitly controlled geometry. In these models, the objective is for the trained network to predict missing representations from only partial input observations. To practically enforce this desired geometry, the authors introduced the Sketched Isotropic Gaussian Regularization (SIGReg) functional, a regularizer that encourages the learned embeddings to have approximately isotropic Gaussian statistics in the latent space, enabling its practical realization.

Although JEPA/LeJEPA focus on self-supervision rather than classification, they emphasize the key role of representation distribution and geometry in the quality of downstream tasks. The LeJEPA formulation is closely related to \rev{autoencoders}~\cite{Hinton-2006-ReducingDimensionality} \rev{and, in particular, to} variational autoencoders (VAE). In \rev{the classical VAE works of Kingma and Welling}~\cite{Kingma-2014-VAE} \rev{and Rezende et al.}~\cite{Rezende-2014-StochasticBackprop}, the encoder $q_\phi(z \mid x)$ is viewed as an optimal (in the variational sense) approximate posterior over the latent variables, chosen to minimize the variation of the likelihood. The decoder $p_\theta(x \mid z)$ explicitly models the output regression. In LeJEPA, the situation is conceptually similar: the encoder $h_\theta(x)$ is trained to provide an optimal (in the sense of the chosen risk functional) encoding of observations in the latent space, yet an explicit decoder is absent, and the quality of representations is evaluated by the predictability of embeddings and their suitability for downstream tasks~\cite{Balestriero-2025-LeJEPA}.

The emergence of a meaningful ``cluster geometry'' in learned representations is not unique to neural feature spaces: multilevel analysis of complex self-organized dynamic systems often introduces an intermediate (meso) level where the effective state is described by clusters/coalitions rather than individual agents, yielding a reduced description with emergent cluster structure \cite{Erofeeva2024Multilevel}.

If the downstream task is classification, then a classifier $p_\psi(y \mid h_\theta(x))$ is built on top of the latent space, and the composition $x \mapsto h_\theta(x) \mapsto
p_\psi(y \mid h_\theta(x))$ becomes a usual classifier. In this sense, the LeJEPA encoder effectively plays the role of a preliminary classifier (feature recognizer), and the results on viewing a neural classifier as an FEC code decoder naturally transfer to this scenario: the latent code $h_\theta(x)$ can be interpreted as a codeword formed by an optimal (in the variational/JEPA sense) encoder, and the subsequent classification decision, correspondingly, as ML decoding in the presence of noise and quantization.

In this study, we significantly advance the view of a neural classifier as an FEC decoder, demonstrating that this interpretation can be formalized in a rigorous theoretical manner. The work provides a unifying framework that complements and integrates the previously mentioned existing research streams on \rev{ECOC}~\cite{Dietterich-Bakiri-1995-ECOC}, \rev{NC}~\cite{papyan2020neuralcollapse}, and JEPA/LeJEPA~\cite{Balestriero-2025-LeJEPA}. The representation geometries consistently reported in the literature, specifically, the approximately isotropic Gaussian clouds observed in LeJEPA and the simplex-like configurations characterizing Neural Collapse, are shown to naturally align with the geometric requirements of classical coding theory for maximizing inter-codeword distances and achieving error-resilience.

The approach differs fundamentally from classical methodologies. Specifically, we consider the entire network, rather than just the output layer, as a sequence of noisy communication channels operating over the (potentially discrete) feature space. We employ the random coding model framework, where class representations are treated as codewords sampled from a codebook rather than as fixed hand-designed labels, drawing inspiration directly from Shannon's \rev{foundational work}~\cite{shannon1949}.

Unlike prior studies \rev{of Neural Collapse}~\cite{papyan2020neuralcollapse} \rev{and LeJEPA}~\cite{Balestriero-2025-LeJEPA}, which primarily observe or postulate this beneficial geometry, our contribution lies in offering a robust theoretical account for its emergence. This account is grounded in the formal requirements for robustness against additive noise and quantization and leverages analogues of Shannon's limit theorems applied to neural architectures. From this coding-theoretic perspective, standard architectural and optimization choices, such as the use of the softmax function,  the cross-entropy loss, and smoothed activation functions, acquire a natural interpretation as practical, albeit approximate, implementations of FEC decoding principles.

Furthermore, these principles extend directly to self-supervised models like LeJEPA, where the generated latent representations serve as codewords that enable highly reliable downstream decoding. This unified viewpoint establishes a framework for the principled design and analysis of both classifiers and generative models, enabling optimization with respect to critical factors such as representation geometry, noise immunity, and energy efficiency.

This work also provides empirical evidence\footnote{Source code for the experiments: \url{https://github.com/iceslice0/fec-experiment}.} that standard neural classifiers can be interpreted as discrete, channel-based systems, extending insights from Neural Collapse beyond geometric convergence. Synthetic experiments demonstrate that cross-entropy training recovers decoding behavior consistent with maximum-likelihood inference over noisy channels, while learned pre-FC representations act as error-correcting codewords.

In addition to the theoretical decoding perspective, we include qualitative
``semantic inversion'' experiments (Fig.~\ref{fig:regnet-recon},~\ref{fig:regnet-recon-ext}) that reconstruct ImageNet~\cite{Russakovsky2015ILSVRC} images from
the latent representation of the deep RegNet model~\cite{Radosavovic2020DesigningRegNet} by matching representations via KL/Jeffreys
divergence under a natural-image total variation (TV) prior. The resulting reconstructions preserve class semantics while varying at
the pixel level, providing an operational illustration of the ``appearance vs. form'' interpretation discussed in
Sec.~\ref{sec:coding-class}. From the self-supervised (JEPA/VAE) viewpoint, this inversion can be interpreted as an optimization-based,
non-amortized decoder for a frozen encoder (the pretrained RegNet trunk), mapping a target latent codeword back to a plausible natural
image under a weak physical prior.

The rest of the paper is organized as follows. Our analysis proceeds
by first formalizing the channel coding perspective on supervised classification in Section 2. Here, we introduce the concepts of random codebooks and
isotropic feature-space noise to frame the problem as maximum-likelihood decoding. Following this foundational work, Section 3 bridges classical coding
theory and modern deep learning by establishing that standard divergence-based losses, such as Kullback-Leibler and Jeffreys divergences, act as smooth
surrogates for Hamming distance. In Section 4, we leverage these insights
to prove that standard softmax classifiers effectively perform maximum-likelihood decoding on learned codewords, linking practical implementations
to optimal decoder theory. Section 5 empirically validates these theoretical
claims. Section 6 concludes the presented study by discussing the practical implications of this framework for future architecture design and robust representation learning.
 
\section{Coding and Classification}\label{sec:coding-class}

Formally, we define the set of \(M\) discrete classes as $\mathcal{X} = \{x_1, x_2, \dots, x_M\}$, and let $\mathcal{Y}$ be the set of all possible observed signals or measurements. The fundamental classification task is to determine the originating class $x \in \mathcal{X}$ given the received signal $y \in \mathcal{Y}$. The optimal classification rule under the Bayesian framework is the Maximum a Posteriori (MAP) decision, provided by:
\begin{equation*}
\hat{x} = \arg\max_{x \in \mathcal{X}} p(x \mid y),
\end{equation*}
where $\hat{x}$ is the predicted class, and $p(x | y)$ is the posterior probability of class $x$ given the observation $y$. In a practical deep learning setup, $p(x|y)$ is approximated by a function (e.~g., a neural network) $\hat{p}(x|y)$ trained on empirical data.
Assume that the classes are balanced, meaning the prior probability of each class is uniform, $p(x) = 1/M$ for all $x \in \mathcal{X}$. By applying Bayes' theorem, $p(x|y) = {p(y|x)p(x)}/{p(y)}$, and acknowledging that the denominator $p(y)$ and the prior $p(x)$ are constants with respect to the $\arg\max$ operation, the MAP rule simplifies to the Maximum Likelihood (ML) rule:
\begin{equation}
\label{MAP=ML}
\hat{x}=\arg\max_{x\in\mathcal{X}} p(x \mid y)= \arg \max_{x\in\mathcal{X}} p(y \mid x),
\end{equation}
\rev{that is, the class} $x$ \rev{most likely to have produced the observed signal} $y$.

This entire classification framework can be directly recast as a decoding problem within the context of information theory. Specifically, the problem is modeled as data transmission over a noisy channel, where the originating class $x \in \mathcal{X}$ is the transmitted \textit{message}, and the observed signal $y \in \mathcal{Y}$ is the \textit{received noisy signal}. The neural classifier, in this view, functions as the optimal decoder implementing the ML rule (\ref{MAP=ML}).

Let $\mathcal U$ be a fixed codeword (feature) space, and let $\mathcal{P}(\mathcal{U})$
denote the set of all probability distributions on $\mathcal U$.
\textit{A forward error-correcting code (or FEC code)} is defined as a pair of mappings $(\mathcal{E}, \mathcal{D})$, where
$\mathcal{E} \colon \mathcal{X} \to \mathcal{P}(\mathcal{U})$ is an \textit{encoder} that assigns to each message
$x \in \mathcal{X}$ a probability distribution of codewords, and $\mathcal{D} \colon \mathcal{Y} \to \mathcal{X}$ is
a \textit{decoder} that reconstructs an estimate $\hat{x} = \mathcal{D}(y)$ of the original message from the observed signal
$y \in \mathcal{Y}$.
We write $c_x \sim \mathcal{E}(\cdot\mid x)$ for a codeword drawn from the encoder distribution of class $x$.
The per-class codebook is $C(x):=\operatorname{supp} E(\cdot\mid x)\subset \mathcal U$, where
$\operatorname{supp}$ denotes the support of a distribution (the set of points assigned nonzero probability).

The set $\overline{\mathcal{C}} := \bigcup_{\mathcal{X}}\mathcal{C}(x)$
is called the \textit{global codebook} or \textit{codebook}.
If the cardinality satisfies \(|\mathcal{C}(x)|=1\) for all \(x\), then the definition coincides with the traditional deterministic coding scheme.

The decoding task consists of recovering the original message $x \in \mathcal{X}$ from the observed signal (received message) $y \in \mathcal{Y}$, which results from the transmission of a codeword through a noisy channel. A \textit{channel} is defined as the triple $(\overline{\mathcal{C}}, p(y  \mid  c), \mathcal{Y})$,
where $p(y  \mid  c)$ is the conditional probability of observing the message $y \in \mathcal{Y}$ given that the codeword $c \in \overline{\mathcal{C}}$ was transmitted.

The process is illustrated in the following Fig.~\ref{fig:channel}.
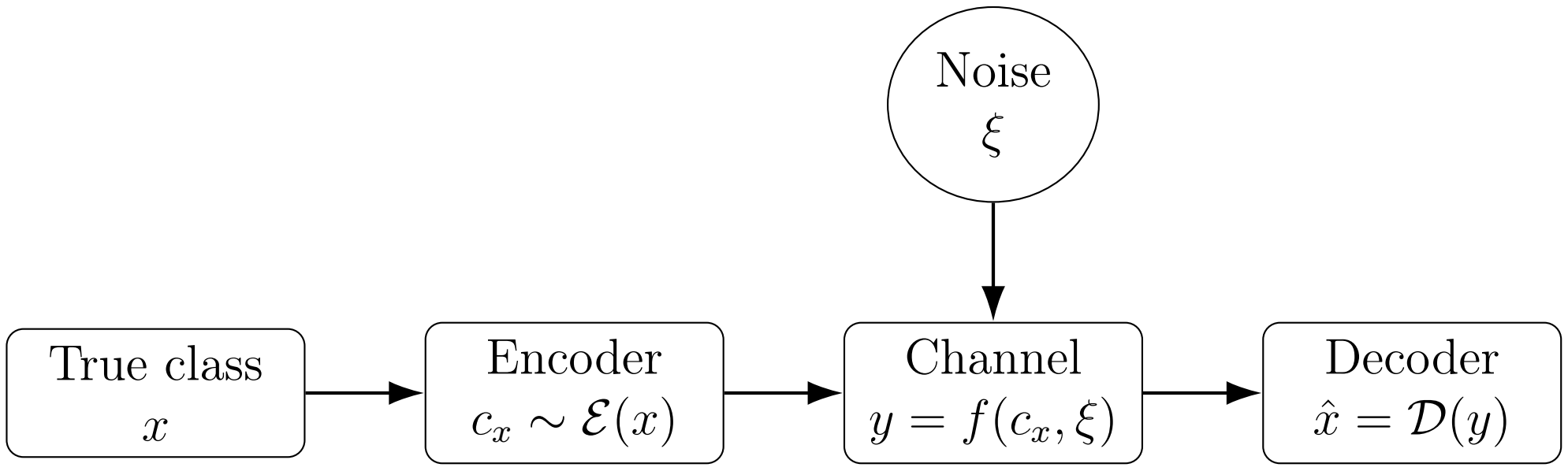
\begin{figure}[t]
    \centering
    \begin{tikzpicture}[node distance=2.5cm, auto,
    box/.style={draw, rectangle, rounded corners, minimum width=2.5cm,
    minimum height=1cm, align=center}, noise/.style={draw, ellipse,
    minimum width=1.5cm, minimum height=1cm, align=center},
    arrow/.style={-{Latex[length=3mm,width=2mm]}, thick}]

\node[box] (input) {True class \\ $x$};
    \node[box, right=1cm of input] (encoder) {Encoder \\ $c_x\sim\mathcal{E}(x)$};
    \node[box, right=1cm of encoder] (channel) {Channel\\ $y=f(c_x,\xi)$};
    \node[box, right=1cm of channel] (class) {Decoder \\ $\hat{x}=\mathcal{D}(y)$};
    \node[noise, above=1cm of channel] (noise) {Noise \\ $\xi$};

\draw[arrow] (input) -- (encoder);
    \draw[arrow] (encoder) -- (channel);
    \draw[arrow] (channel) -- (class);
    \draw[arrow] (noise) -- (channel);

\end{tikzpicture}
     \caption{Classification as a transmission model}
    \label{fig:channel}
\end{figure}

We formally introduce a set $\Xi$ to encompass the hidden parameters that characterize the influence of noise on the system. The formation of the observable signal $y \in \mathcal{Y}$ from the originating class $x \in \mathcal{X}$ is described by the deterministic function $f : \overline{\mathcal{C}} \times \Xi
\to \mathcal{Y}$, according to the relationship:
$$y = f(c, \xi).$$

In this model for the deterministic case \(|\mathcal{C}(x)|=1\), the noise variable $\xi \in \Xi$ is assumed to follow a known and specified distribution, denoted by $\xi \sim p_\Xi(\xi)$, allowing the conditional probability $p(y \mid c_x) = p(y \mid  x)$, which defines the communication channel, to be entirely derived from the deterministic mapping $f$ and the known noise distribution $p_{\Xi}(\xi)$:
$$
 p(y \mid c_x)
= \int\limits_{\Xi} \delta\bigl(y - f(c_x, \xi)\bigr) \, p_{\Xi}(\xi) \, d\xi,
$$
where
$\delta(\cdot)$ is the Dirac delta function reflecting the deterministic
dependence of $y$ on $x$ and $\xi$.

Thus, the classification problem reduces to decoding a message transmitted through a noisy channel with characteristic $p(y \mid c)$.

The formal analogy between supervised classification and error-correcting decoding allows for a deeper conceptual interpretation grounded in Plato's theory of Forms. In practical machine-learning settings, the only directly observable quantity is the signal $y$. The underlying class label $x$, the characteristics of the implicit ``channel'', and the noise $\xi$ are not directly measurable; they are latent variables that must be inferred. This parallels the Platonic distinction between ideal essences and their imperfect manifestations. Within this framework, we interpret the class $x$ as a Platonic Form. For example, the class ``table'' corresponds to the ideal and immutable Form of Tableness, which is, in turn, an abstract essence that defines what it means to be a table. Any specific observation $y$ (e.~g., an image of a particular table) is therefore not the Form itself but an approximation, an imperfect copy, or a specific instance shaped by contingent physical conditions.

The noise term $\xi$ captures both (i) benign intra-class variability (differences in viewpoint, color, or material that preserve recognizability) and (ii) adverse perturbations (occlusion, cropping, illumination changes, sensor noise) that distort the observation in ways that obscure its relation to the underlying Form. In this sense, noise embodies the gap between the ideal entity and its imperfect expression in the empirical world.

Consequently, the classification-as-communication model (Fig.~\ref{fig:channel}) can be viewed as an operational, mathematical rendering of Platonic idealism. The learning system's objective is to reconstruct the ideal class $x$ from a corrupted observation $y$, that is, to recover the underlying Form from its noisy physical instantiation. An optimal classifier serves as a decoder that most faithfully infers the ideal entity by compensating for the distortions introduced by the ``worldly'' noise. This perspective provides an interpretive bridge between classical philosophical notions of abstraction and contemporary approaches to robust geometric representation learning.

\begin{figure}[t]
    \centering
    \begin{tikzpicture}[node distance=1.5cm]

\node (message) [draw, rectangle, rounded corners, align=center] {True class \(x\)};
\node (encoder) [draw, rectangle, rounded corners, align=center, below=of message] {Random coding \(c \in \mathcal{C}(x)\)};
\node (channel) [draw, rectangle, rounded corners, align=center, below=of encoder] {Noisy channel \(y = f(c,\xi)\)};
\node (decoder) [draw, rectangle, rounded corners, align=center, below=of channel] {ML decoding \(\hat{x}=\arg\max\limits_{x\in\mathcal{X}} \max\limits_{c\in\mathcal{C}(x)} p(y \mid c)\)};
\node (output) [draw, rectangle, rounded corners, align=center, below=of decoder] {Recognized class \(\hat{x}\)};

\draw[->] (message) -- node[right]{Instantiation} (encoder);
\draw[->] (encoder) -- node[right]{Realization} (channel);
\draw[->] (channel) -- node[right]{Observation} (decoder);
\draw[->] (decoder) -- node[right]{Classification} (output);

\end{tikzpicture}
     \caption{Classification as random coding with forward error correction}
    \label{fig:random_code}
\end{figure}
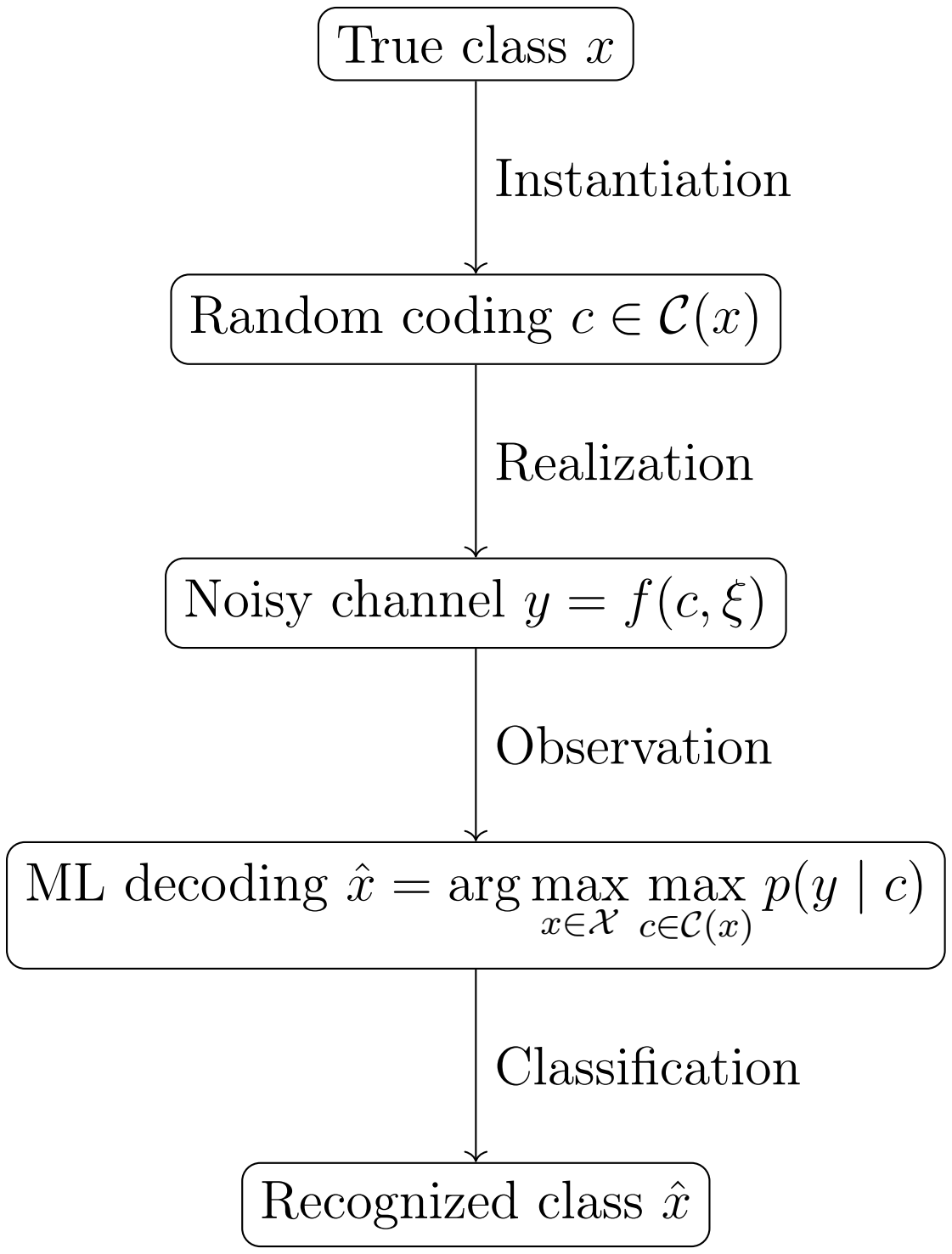

However, this approach is still not practical because it provides no insight into the codeword space. Note that a single class, for example, the class of tables, can have many features, such as color, size, and position, as well as many subclasses, such as coffee table, dining table, and writing desk. The space of binary features can be considered, in which a single class corresponds to multiple codewords chosen at random. A random codebook is therefore understood as a collection of class-dependent binary feature vectors sampled or learned so that different classes remain well separated in Hamming distance. For simplicity, assume that the number of codewords for each class is the same. The considered scheme is shown in Fig.~\ref{fig:random_code}.

Here, in contrast to the previous scheme shown in Fig.~\ref{fig:channel}, each class \(x \in \mathcal{X}\) does not have a single codeword, but rather a codebook \(\mathcal{C}(x)\) of codewords of length $n$, which are hard binary vectors
$c\in\mathcal{U} :=\{0,1\}^n$.
\begin{equation}
\label{|C(x)|} |\mathcal{C}(x)| = K \quad \forall\, x \in \mathcal{X}.
\end{equation}
The channel output is a vector of soft bits, i.~e., $y\in \mathcal{Y}:=[0,1]^n$, and $y_i$ is interpreted as
the posterior bit probability $y_i=\mathbb{P}(b_i=1\mid\cdot)$.

Assume that, under random coding, the choice of a codeword \(c \in \mathcal{C}(x)\)
is uniform, i.~e.
\begin{equation}
    \label{p(c|x)}
p(c \mid x)=\frac{1}{K},
\end{equation}
and the observed value \(y\) is formed according to the formula \(y = f(c, \xi)\), where
\(\xi\) is random noise, and the function \( f : \{0,1\}^n \times \Xi\to \mathcal{Y} \)
describes the action of the noisy channel.

Then one can assert that the ML classification problem under consideration is equivalent to the problem of (ML) decoding of FEC codes with random coding.
Indeed, given that under uniform random coding equality (\ref{p(c|x)})
holds for all \(c \in \mathcal{C}(x)\), the conditional probability of the
observation \(y\) for a given class \(x\) is defined as
\begin{equation*}
p(y \mid x)=\sum_{c \in \mathcal{C}(x)} p(y \mid c)\,p(c \mid x)
=\frac{1}{K}\sum_{c \in \mathcal{C}(x)} p(y \mid c).  \end{equation*}

Using the obtained representation, the equality (\ref{MAP=ML}) can be expressed in the considered situation as
$$
\hat{x}=\arg\max_{x\in\mathcal{X}} \sum_{c \in \mathcal{C}(x)} p(y \mid c).
$$

Thus, the problem of recovering the original class $x$ from the observed
signal $y$ reduces to choosing the class for which the total probability of
obtaining $y$ over all codewords associated with that class is maximal. This
corresponds to the ML decoding problem of an FEC code composed of the union
of codebooks $\mathcal{C}(x)$ for all $x \in \mathcal{X}$ under the uniform random coding assumption.

Adopting the robust coding-theoretic interpretation of neural classifiers provides several critical insights necessary for advancing machine learning theory:
\begin{itemize}
\item \textit{Modeling the Complete Distribution, Not Just Prototypes}: Contrary to the idea that a neural network converges on a single ``ideal'' prototype of a class (a static Platonic Form), the network, when viewed as an optimal decoder, must effectively model the entire conditional probability distribution $p(y \mid x)$ across all possible realizations of the class $x$. The goal is not merely to find the mean, but to understand the entire manifold of feature vectors belonging to that class.

\item  \textit{Data Diversity Expands the Codebook}: The presence of high data diversity is crucial, as it directly corresponds to effectively expanding the codebook ($\bar{\mathcal{C}}$) of the FEC scheme. A larger, more richly sampled codebook provides the decoder with more distinct class-specific encodings, thereby improving the classifier's generalization capacity and robustness.

\item \textit{Robustness to Noise through Codebook Density}: The system's robustness to noise stems from the geometric arrangement of the codebook. Suppose the set of codewords (class feature distributions) densely covers the space $\mathcal{Y}$. In that case, the addition of noise will rarely shift the observed signal $y$ into the decision region belonging to another class $\hat{x} \neq x$. This noise immunity is directly proportional to the effective minimum distance between the codewords.

\item \textit{Connection to Shannon's Limit Theorems}: This framework establishes a direct link to Shannon's channel coding theory. Specifically, as the effective code length (often interpreted as the feature vector dimensionality or depth) increases, the probability of a decoding error (misclassification) is predicted to decay exponentially, provided the rate remains below the channel capacity. This explains the performance gain observed in deeper or wider networks.

\item \textit{Connection to Neural Collapse (NC)}: The geometric regularity observed during NC is the practical manifestation of the decoder optimizing its codebook geometry. NC forces the features of a class to collapse toward their mean in a simplex-like or isotropic Gaussian arrangement. This configuration is the optimal codebook geometry required by classical coding theory to maximize the minimum distance between codewords, thereby directly improving noise immunity and simplifying decoding.
\end{itemize}

We distinguish two notions:

\begin{enumerate}
    \item FEC Shannon-optimality (optimality). For any $K\ge 1$, we call the stochastic code \textit{optimal} if the codebook
    $\overline{\mathcal{C}}$ with the ML decoder is an optimal error-correcting code in the classical Shannon sense for the
    specified channel. For $K=1$, this reduces to the standard ``one codeword per class'' coding setup.
    \item Dominance. Independently of Shannon-optimality, we say that the per-class codebook $\mathcal{C}(x)$ is
    dominant (for a given observation $y$) if there exists a codeword
    \begin{equation}\label{c*}
    c^*(x)=\arg\max_{c\in\mathcal{C}(x)} p(y \mid c),
    \end{equation}
    such that
    \[
    p(y\mid c) \le \varepsilon\, p(y\mid c^*(x)), \qquad \varepsilon \ll 1, \qquad \forall\,c\in\mathcal{C}(x),\ c\neq c^*(x).
    \]
    That is, the likelihood mass within $\mathcal{C}(x)$ is concentrated on a single codeword, so the mixture over $K$ codewords
    per class is well-approximated by its maximum in the dominant-codeword regime. Thus, codeword dominance means that one candidate representation explains the observation so much better than the other same-class codewords that the class likelihood can be replaced by its largest term.
\end{enumerate}

\begin{theorem}\label{th:optimal_fec} Suppose
that \(C\) has the optimal union codebook and the dominance condition~\eqref{c*} holds.
Then the following equivalence holds:
$$\arg\max_{x\in\mathcal{X}} \sum_{c \in
\mathcal{C}(x)} p(y \mid c)=\arg\max_{x\in\mathcal{X}} \max_{c \in \mathcal{C}(x)}
p(y \mid c).  $$
\end{theorem}

\begin{proof} For each class \(x\in\mathcal{X}\), let $c^*(x)$ denote the dominant codeword defined in (\ref{c*}). Introduce $$C^{-}(x)
:= C(x) \setminus\{c^*(x)\}.$$

The sum of likelihoods over the codebook
$C(x)$ is \begin{equation} \label{S(x)} S(x) :=\sum_{c
\in \mathcal{C}(x)} p(y \mid c)=p(y \mid c^*(x))+\sum_{c\in\mathcal{C^-}(x)} p(y \mid c).
\end{equation}

Assuming the dominance property~\eqref{c*}, for any non-dominant codeword \(c\in\mathcal{C^-}(x)\),
$$
p(y \mid c)\le \varepsilon\, p(y \mid c^*(x)),
$$
for some sufficiently small $\varepsilon > 0$.

Assumption (\ref{|C(x)|}) states that each codebook contains $K$ codewords. Hence,
$$
\sum_{c\in\mathcal{C}^-(x)} p(y \mid c) \le (K-1)\varepsilon\, p(y \mid c^*(x)).
$$

Substituting into (\ref{S(x)}) gives
$$
p(y \mid c^*(x)) \le S(x)
\le p(y \mid c^*(x))\bigl(1+(K-1)\varepsilon\bigr).
$$

When $\varepsilon$ is sufficiently small (e.~g., $\varepsilon
\ll (K-1)^{-1}$), the multiplier $(1+(K-1)\varepsilon)$ is close
to 1; thus $S(x)$ is dominated by the contribution of \(c^*(x)\). Therefore,
$$
\hat{x}=\arg\max_{x\in\mathcal{X}} S(x)=\arg\max_{x\in\mathcal{X}}
\Bigl(p(y \mid c^*(x))\bigl(1+(K-1)\varepsilon\bigr)\Bigr).
$$

Since the factor $(1+(K-1)\varepsilon)$ does
not depend on $x$, the \(\arg\max\) preserves ordering:
$$
\hat{x}=\arg\max_{x\in\mathcal{X}} p(y \mid c^*(x))=\arg\max_{x\in\mathcal{X}}
\max_{c\in\mathcal{C}(x)} p(y \mid c).
$$

This completes the proof.
\end{proof}

Thus, it is formally demonstrated that, under the condition of optimal dominant FEC, selecting the class based on the total likelihood is mathematically equivalent to choosing the class with the maximal likelihood of an individual codeword within that class. This confirms that the ML rule provides the optimal means for decoding the class information from the observed, noisy signal.

\rev{Noise in the feature space is called} \textit{isotropic}, i.~e., uniform in all directions and uncorrelated across coordinates, if the likelihood function
$p(y \mid c)$ depends solely on the distance between the observation
$y$ and the codeword $c$. In this case,
\begin{equation}
\label{g||}
    p(y \mid c) = g\bigl(d(y, c)\bigr),
\end{equation} where \(d(\cdot,\cdot)\) is a metric (distance function), and \(g:\mathbb{R}_{\ge0} \to \mathbb{R}_{\ge0}\) is a monotonically decreasing function with sufficiently rapid decay.
\rev{We say that} $g$ has a \textit{multiplicative decay envelope} if there exists a nonincreasing function
$\rho:\mathbb{R}_{\ge 0}\to(0,1]$ such that for all $r,\vartriangle\ge 0$,
\begin{equation}\label{eq:rho}
\frac{g(r+\vartriangle)}{g(r)}\le \rho(\vartriangle).
\end{equation}

The following theorem gives a sufficient condition under which the optimal FEC code \(\overline{\mathcal{C}}\) is also dominant under random coding with isotropic noise.

\begin{theorem}[Dominance from metric noise and code separation]\label{th:dominance_metric}
Assume a channel with isotropic noise, where $g$ has a multiplicative decay envelope.
Fix a class $x$ and its codebook $\mathcal{C}(x)$, and let its minimum distance be
$d_{\min}(x)=\min_{c\neq c'\in\mathcal{C}(x)} d(c,c')$.
If an observation $y$ satisfies the condition
$$d(y,c_0)\le r<\tfrac{1}{2}d_{\min}(x)$$ for some $c_0\in\mathcal{C}(x)$, then
$c^*(x)=c_0$ in~\eqref{c*} and for every $c\in\mathcal{C}(x)\setminus\{c_0\}$,
\begin{equation}\label{eq:dominance_eps}
p(y\mid c)\le \varepsilon\,p(y\mid c_0),
\end{equation}
where $
\varepsilon:=\rho\left(d_{\min}(x)-2r\right).
$
\end{theorem}

\begin{proof}
For any $c\neq c_0$, by the triangle inequality
$d(y,c)\ge d(c_0,c)-d(y,c_0)\ge d_{\min}(x)-r$, hence
$d(y,c)-d(y,c_0)\ge d_{\min}(x)-2r=: \vartriangle>0$.
Since $g$ is decreasing and~\eqref{eq:rho} holds,
\[
\frac{p(y\mid c)}{p(y\mid c_0)}
=\frac{g(d(y,c))}{g(d(y,c_0))}
\le \frac{g(d(y,c_0)+\vartriangle)}{g(d(y,c_0))}
\le \rho(\vartriangle),
\]
which gives~\eqref{eq:dominance_eps}. The condition $r<\tfrac{1}{2}d_{\min}(x)$ implies $\vartriangle>0$, so $c_0$ is the unique maximizer.
\end{proof}

\begin{corollary}[Theorem~\ref{th:dominance_metric} $\Rightarrow$ Theorem~1 regime]\label{cor:th2_to_th1}
Under the assumptions of Theorem~\ref{th:dominance_metric}, whenever $d(y,c_0)\le r<\tfrac{1}{2}d_{\min}(x)$ holds for a class $x$,
the dominance condition of Theorem~1 holds for that class with $\varepsilon=\rho(d_{\min}(x)-2r)$.
Consequently, in this low-noise (bounded-metric) regime, the class likelihood
$S(x)=\sum_{c\in\mathcal{C}(x)}p(y\mid c)$ is well-approximated by its dominant term $p(y\mid c^*(x))$ as in Theorem~1.
\end{corollary}

In essence, the theorems demonstrate that for a neural network to achieve effective quantization and noise robustness, its feature space must exhibit metric regularity, ensuring that small perturbations from numerical rounding or input corruption result in minimal changes to the representation. Additionally, maximizing the separation of class-dependent codewords creates a geometry with large, uniform inter-class distances. This arrangement, similar to structures observed in NC and predicted by coding theory, helps prevent quantization errors and isotropic noise from crossing decision boundaries. In essence, a well-quantizing network resembles an optimal error-correcting code, with robust decoding resulting from both metric regularity and the distance between codewords.

Furthermore, one can conclude that the classification problem is analogous to the conventional challenge of optimal coding for a noisy channel, assuming that isotropic observation noise exists within the feature space. This assumption is \rev{reasonable as long as only reliable features that are not prone to noise interference are used.}
 
\section{Distance and Divergence for a Binary Channel}

In coding theory, the primary measure of code quality and error resilience is the Hamming distance, which counts the number of positions in which two binary sequences differ. In contrast, classification models, particularly neural networks trained with probabilistic outputs, are typically optimized using the Kullback-Leibler divergence $D_{\mathrm{KL}}$, a widely used measure of dissimilarity between the predicted and true class distributions. If classification is to be interpreted as a form of decoding, as posited in our unified framework, then these two \rev{measures} cannot remain conceptually isolated. Instead, a formal correspondence between the geometric measure used in coding and the information-theoretic measure used in classification must exist. Intuitively, if both processes address the same underlying task of recovering latent structure from corrupted observations, then the respective error measures should admit a mathematical equivalence.

The analysis that follows establishes this equivalence rigorously. It demonstrates how the separation between codewords in the coding space, as measured by the Hamming distance, aligns with divergence-based loss functions in neural classifiers. This connection enables classical results from coding theory to be directly transferred to the study and design of modern neural networks.

An asymmetric memoryless binary channel can be characterized by the error probabilities \( p_0 \) for the transition \( 0 \to 1 \) and \( p_1 \)
for the transition \( 1 \to 0 \). For each bit, the true binary value \(
b \in \{0,1\} \) specifies the ``soft'' label \( q = (\mathbb{P}(\text{observe } 0), \mathbb{P}(\text{observe } 1))\) as follows: \[ q =
\begin{cases} (1-p_0,\, p_0), & \text{if } b=0,\\[1mm] (p_1,\, 1-p_1), &
\text{if } b=1.  \end{cases} \]

To develop a unified framework that connects coding-theoretic and probabilistic notions of error, we extend the classical Hamming distance ($d_H$) to operate on probability distributions by replacing it with the \(L_1\) distance. This formulation is constructed so that it coincides with the standard Hamming distance when applied to hard codewords. Let $P$ and $Q$ denote the vectors of bit-wise marginals of two distributions
on the binary alphabet $\{0,1\}$ for a length-$n$ codeword, i.~e.,
$P_i=\mathbb{P}(b_i=1)$ and $Q_i=\mathbb{P}(b_i=1)$.
We define their soft Hamming distance as
$$
d_H(P,Q) \;:=\; \sum_{i=1}^n |P_i-Q_i|.
$$
In the probabilistic setting, this expression corresponds to the $L_1$ distance between the vectors of bitwise marginals. When $P,Q\in\{0,1\}^n$ are hard codewords, this reduces to the standard Hamming distance.
This distance will be used for both ``hard'' and ``soft'' labels and will be referred to as \rev{the} Hamming distance for brevity.

\rev{We introduce} the Jeffreys divergence $D_J$, which is the symmetric Kullback-Leibler divergence
between probability distributions $P$ and $Q$ on the set of classes $\mathcal{X}$:
$$
D_J(P\| Q) := D_{\mathrm{KL}}(P\|Q) + D_{\mathrm{KL}}(Q\|P),
$$
$$
D_{\mathrm{KL}}(P\Vert
Q) := \sum_{x \in \mathcal{X}} P(x)\,\ln\frac{P(x)}{Q(x)},
$$
and the cross-entropy
$$
H(P, Q) := H(P) + D_{\mathrm{KL}}(P\Vert Q),
$$
where
\(H(P)\) is the entropy of the distribution \(P\).
For one-hot labels, this measure coincides with the negative log-likelihood, which underlies the widespread use of cross-entropy in machine learning.

Let $\varphi: \Omega \subset\mathbb{R}^n \to \mathbb{R}$ be a differentiable, strictly convex function.
Then the Bregman divergence generated by $\varphi$ is defined as
$$
D_{\varphi}(x, y) := \varphi(x) - \varphi(y) - \langle \nabla \varphi(y),\, x - y \rangle.
$$

Note that the Bregman divergence generated by
$\varphi(p) = \sum_{i=1}^n p_i \log p_i$ coincides with the Kullback-Leibler divergence $D_{\varphi}(x, y)  = D_{\mathrm{KL}}(x \Vert y)$.

For an optimal maximum-likelihood (ML) softmax classifier, the predicted value $\hat{b}$ determines the corresponding output distribution
$$
P = \begin{cases} (1-p_0,\, p_0), & \text{if }
\hat{b}=0,\\[1mm] (p_1,\, 1-p_1), & \text{if } \hat{b}=1.  \end{cases}
$$

For each individual bit, two cases arise:
\begin{enumerate}
    \item $\mathbf{No \; error \; (\hat{b}=b).}$

    In this case, the true distribution $q$ coincides with the classifier output $P$. Consequently,
$D_{\mathrm{KL}}(q\|P) = D_{\mathrm{KL}}(P\|q)=0$ and therefore the Jeffreys divergence is
    \begin{equation}
    \label{Dsym0} D_J(q\|P)=0.
    \end{equation}

    \item $\mathbf{Error \; (\hat{b}\neq b).}$

    The Jeffreys divergence takes a fixed positive value
    \begin{equation}
    \label{Delta}
    D_J(q\|P) = \Delta,
    \end{equation} where
    \begin{equation} \label{DeltaDelta} \Delta := D_{\mathrm{KL}}\Bigl((1-p_0,
    p_0) \,\big\|\, (p_1, 1-p_1)\Bigr) + D_{\mathrm{KL}}\Bigl((p_1, 1-p_1)
    \,\big\|\, (1-p_0, p_0)\Bigr) \end{equation} for the case \( b=0 \)
    (the case \( b=1 \) yields the same constant).
\end{enumerate}

Thus, every bit error contributes exactly $\Delta$ to the total divergence.
For a binary vector of length $n$, the total Jeffreys divergence between the \rev{true} bit sequence $b = (b_i)$ and the classifier output $\hat{b} = (\hat{b}_i)$ is therefore
$$
D_J^{\mathrm{total}} := \sum\limits_{i: \, b_i \neq \hat{b}_i}
\Delta  = \Delta \, d_H(b,\hat{b}),
$$
where \( d_H(b,\hat{b}) \) denotes the Hamming distance between the actual and predicted bit vectors. Since divergence accumulates only over error positions, the Jeffreys divergence is directly proportional to the number of bit errors.

\begin{theorem}
\label{th:jeffreys-hamming}
Minimizing Jeffreys divergence between the true soft
labels and the output of the optimal softmax classifier is equivalent to
minimizing the Hamming distance between the observed and true binary features.
\end{theorem}

\begin{proof} When considering each individual bit, there are two cases:
either an error occurred, or it did not.

If no error occurred, then the true value \( b \) coincides with the observed
value \(\hat{b}\). The true distribution \(q\) and the classifier output \(P\)
coincide and therefore, according to (\ref{Dsym0}), \( D_J(q\|P)=0
\).

If an error occurs, the observed value \(\hat{b}\) differs from the true
value \( b \). Let \( b=0 \) (the case \( b=1 \) is treated similarly). Then
the true distribution is \[ q = (1-p_0,\, p_0), \] and the observed distribution,
owing to the error, corresponds to \(\hat{b}=1\) and \[ P = (p_1,\, 1-p_1).  \]

Therefore, according to (\ref{Delta}) and (\ref{DeltaDelta}), the
Jeffreys divergence equals \[ \Delta = (1-p_0)\ln\frac{1-p_0}{p_1} + p_0
\ln\frac{p_0}{1-p_1} + p_1\ln\frac{p_1}{1-p_0} + (1-p_1) \ln\frac{1-p_1}{p_0}.
\]

Note that for fixed \(p_0\) and \(p_1\) the value \(\Delta\) is a constant
that does not depend on the specific bit. Thus, if an error occurred, the
contribution to the total Jeffreys divergence for that bit equals \(\Delta\).

From the two cases considered above, it is clear that when a vector of \(n\)
bits is transmitted, the sum of the Jeffreys divergence over all bits equals
$$
D_J^{\mathrm{total}} = \Delta \cdot \bigl(\text{number of
bits in which an error occurred}\bigr).
$$

\rev{By definition,} the number of erroneous
bits is determined by the Hamming distance \( d_H(b,\hat{b}) \) between the
true vector \( b \) and the observed vector \( \hat{b} \). Consequently, \[
D_J^{\mathrm{total}} = \Delta \,d_H(b,\hat{b}).  \] This proves
that the Jeffreys divergence between the true soft labels and the output
of the optimal softmax classifier is proportional to the Hamming distance
between the observed and true binary features for an asymmetric memoryless
binary channel.  \end{proof}

Note that if the channel is symmetric, the Jeffreys divergence is equivalent
to the (one-sided) Kullback-Leibler divergence. In what follows,
we \rev{consider general asymmetric binary channels, with the symmetric channel as a special case.} Hinton's distillation uses the
(one-sided) Kullback-Leibler divergence, whereas our approach employs the
Jeffreys divergence, which provides advantages during training.

\begin{corollary} Let the memoryless binary channel be symmetric (BSC), that
is, \(p_0 = p_1 = p\). Then the soft labels for each bit take the form
\[
q = \begin{cases}
(1-p,\, p), & \text{if } b=0,\\[1mm]
(p,\, 1-p), & \text{if } b=1,
\end{cases}
\]
and the optimal softmax classifier based on maximum likelihood produces
\[ P = \begin{cases}
(1-p,\, p), & \text{if
} \hat{b}=0,\\[1mm]
(p,\, 1-p), & \text{if } \hat{b}=1.
\end{cases}
\]
In this case, the one-sided Kullback-Leibler divergence is computed as
\[
D_{\mathrm{KL}}(q\|P) = (1-p)\ln\frac{1-p}{p} + p\ln\frac{p}{1-p}
= (1-2p)\ln\frac{1-p}{p},
\]
and the Jeffreys divergence takes the form
\[
D_J(q\|P) = 2\,D_{\mathrm{KL}}(q\|P) = 2\,(1-2p)\ln\frac{1-p}{p}.
\]
In this case, minimizing the Jeffreys divergence
\(D_J(q\|P)\) is equivalent to minimizing the
Kullback-Leibler divergence \(D_{\mathrm{KL}}(q\|P)\). At the same time, for a
binary vector of length \(n\), the total Kullback-Leibler divergence equals \[
D_{\mathrm{KL}}^{\mathrm{total}} = D_{\mathrm{KL}}(q\|P) \, d_H(b,\hat{b}).
\] In other words, minimizing the Kullback-Leibler divergence between the
true soft labels and the output of the softmax classifier is equivalent
to reducing the number of bit errors, which corresponds to minimizing
the Hamming distance between the observed and true binary features under the
BSC assumption.
\end{corollary}

In practice, one might wish to use Jeffreys
divergence rather than the one-sided Kullback-Leibler divergence, but the
former is not directly computable from hard labels because it contains \(\log 0\).
Hard labels can be viewed as rounded soft labels with additional
rounding noise; for realistic datasets, this noise can be large, especially
for simple models, which limits how close the smooth and discrete models can
be made.

A practical way to obtain soft labels is to use a more accurate pretrained
model as a teacher. This approach, proposed in~\cite{hinton2015distilling} under
the name knowledge distillation, can be interpreted in the present
framework as transferring an approximately optimal codebook (and the corresponding
ML decoder) from a complex teacher network to a simpler student
network.
 
\section{Softmax Decoder with a Linear Layer}\label{sec:softmax}

A natural and critical question arises regarding the relationship between the discrete forward-error-correcting coding (FEC) view established in the preceding sections and the widely adopted architecture in practical deep learning: the ``linear layer + softmax + cross-entropy'' classifier.

This section formally addresses this connection. An ``ideal'' reference decoder and its neural implementation are fixed, and it is shown that, under suitable assumptions, the last layer of a standard classifier coincides with the ML decoder of the FEC code. This equivalence is pivotal, as it reveals that the conventional neural classification pipeline inherently performs optimal decoding when viewed through the lens of coding theory, thereby providing a robust theoretical grounding for common architectural choices.

Throughout this section, the following assumptions are used. As in previous sections, $y \in [0,1]^n \subset \mathbb{R}^n$ is a soft binary feature vector, specifically the flattened tensor of penultimate-layer outputs of the neural network, where each coordinate is interpreted as \rev{a} Bayesian confidence.

\textbf{Assumption 1 --- Optimal FEC Decoder:}
\begin{itemize}
    \item 	An optimal codebook exists:
\[
  \overline{\mathcal{C}} = \{c_{i,k} : i=1,\dots,M,\ k=1,\dots,K\}.
\]

\item 	A maximum-likelihood (ML) decoder selects the class whose codeword minimizes the Hamming distance to the input features $d_H(y,c_{i,k})$.

\item 	Existence of such a codebook follows from the Shannon Channel Coding Theorem~\cite{shannon1949}.
\end{itemize}

\textbf{Assumption 2 --- Neural Realization of Distance Computation:}
\begin{itemize}
    \item A neural network can implement (or approximate) the mapping from raw input to per-codeword distances.
    \item For each codeword $c_i$, the Hamming distance $d_H(y,c_i)$ can be written as an affine function of $y$.

    \item The final linear layer can therefore produce logits of the form
\begin{equation}
\label{z_i-lb}
  z_i(y)
  = -\lambda\, d_H(y,c_i) + \beta(y), \quad \lambda>0,
\end{equation}
matching the structure of an ML distance-based decoder.
\end{itemize}

\textbf{Assumption 3 --- Exponential Likelihood Surrogate}
\begin{itemize}
    \item	Class-conditional likelihoods have an isotropic exponential profile
    $$
    p(y\mid c)\propto \exp(-\lambda d_H(y,c)),\quad y\in[0,1]^n,\ c\in\{0,1\}^n.
    $$
    This makes ML decoding equivalent to minimum-Hamming-distance decoding in the dominant-codeword regime.
\end{itemize}

\begin{lemma}[approximation criterion]
\label{lm:cw_dominance}
Let $\{z_{i,k}\}_{k=1}^K\in\mathbb{R}$ be the codeword logits for class $i$, and define
\[
s_i := \log\sum_{k=1}^{K} \exp{z_{i,k}},\qquad a_i := \max_{k} z_{i,k},\qquad
\delta_i := s_i - a_i.
\]
Then $\delta_i\in[0,\log K]$, and
\begin{equation*}\label{eq:logit_margin}
s_i = a_i + \log\!\Big(1+\sum_{k\neq k^*} \exp{z_{i,k}-a_i}\Big),
\quad k^*=\arg\max_k z_{i,k}.
\end{equation*}
In particular, the max-codeword decoder $\arg\max_i a_i$ coincides with the mixture decoder $\arg\max_i s_i$ whenever the max-logit margin satisfies
\begin{equation}\label{eq:cw_dominance}
a_{i^*}-\max_{j\neq i^*} a_j \;>\; \max_{j}\delta_j,
\quad\text{where } i^*=\arg\max_i a_i.
\end{equation}
For $K=1$ we have $\delta_i=0$ and the two decoders coincide exactly.
\end{lemma}
\begin{proof}
Let $a_i=\max_k z_{i,k}$ and choose $k^\ast\in\arg\max_k z_{i,k}$.
Then
\[
s_i=\log\sum_{k=1}^K \exp{z_{i,k}}
=\log\Big(\exp{a_i}\sum_{k=1}^K \exp{z_{i,k}-a_i}\Big)
=a_i+\log\sum_{k=1}^K \exp{z_{i,k}-a_i}.
\]
Since $z_{i,k}-a_i\le 0$ for all $k$, we have $1\le \sum_k \exp{z_{i,k}-a_i}\le K$,
hence $\delta_i=s_i-a_i\in[0,\log K]$, and also
\[
s_i=a_i+\log\!\Big(1+\sum_{k\neq k^\ast} \exp{z_{i,k}-a_i}\Big).
\]

To prove the decoder criterion, let $i^\ast\in\arg\max_i a_i$.
For any $j\neq i^\ast$,
\[
s_j \le a_j+\delta_j \le \max_{l\neq i^\ast} a_l + \max_l \delta_l
< a_{i^\ast} \le s_{i^\ast},
\]
where we used $\delta_{i^\ast}\ge 0$. Therefore $\arg\max_i s_i=\arg\max_i a_i$.

Optionally, if $\gamma_i:=a_i-\max_{k\neq k^\ast} z_{i,k}\ge 0$, then
\[
\sum_{k\neq k^\ast} \exp{z_{i,k}-a_i} \le (K-1)\exp{-\gamma_i},
\qquad
\delta_i \le \log\!\big(1+(K-1)\exp{-\gamma_i}\big).
\]
\end{proof}

Condition~\ref{eq:cw_dominance} holds exactly for one codeword per class \(K=1\) but it is not
asserted to hold universally. In Section~\ref{sec:experiments} we empirically observe that the decoder behavior is consistent with
it for low BSC noise $q$ and small codebook multiplicity $K$, while for larger $q$ and $K$ the assumption breaks,
suggesting that multiple codewords may contribute non-negligibly to the class likelihood.

\begin{corollary}[Theorem~\ref{th:dominance_metric} $\Rightarrow$ approximation]\label{cor:t2_to_lemma1}
Assume the setting of Theorem~2 for a fixed class $i$ and observation $y$,
and define codeword logits (up to an additive constant) by
$z_{i,k} := \log p(y\mid c_{i,k})$.
If Theorem~2 yields $p(y\mid c_{i,k})\le \varepsilon\,p(y\mid c_{i,k^\ast})$
for all $k\neq k^\ast$, then
\[
z_{i,k}-z_{i,k^\ast}\le \log\varepsilon\quad (k\neq k^\ast),
\qquad
\delta_i \le \log\!\big(1+(K-1)\varepsilon\big).
\]
In particular, in the low-noise/separated regime where $(K-1)\varepsilon\ll 1$,
the mixture logit $s_i=\log\sum_k \exp{z_{i,k}}$ satisfies $s_i=a_i+o(1)$,
so Lemma~1 applies with an explicit bound on $\delta_i$.
\end{corollary}

\begin{theorem}[Neural Implementation of the Optimal ML Decoder]\label{th:optimal_linear_sum}
Under Assumptions 1-3, if the condition~\eqref{eq:cw_dominance} holds, then the ML decoder corresponding to the optimal codebook can be realized by a single fully connected linear layer with \(M\) outputs, followed by a softmax activation. In particular, for each class \(i \in \{1,\ldots,M\},\) the contributions of its \(K\) codewords \(\{c_{i,k}\}_{k=1}^K\) can be absorbed into a single pair of ``effective'' parameters \((w_i^{\mathrm{eff}}, b_i^{\mathrm{eff}})\) such that the class logit takes the standard form
$$
z_i(y) = \langle w_i^{\mathrm{eff}}, y\rangle + b_i^{\mathrm{eff}}.
$$
\end{theorem}
\begin{proof}
By Assumption 1, each class \(i\) is associated with a codebook
$\mathcal{C}_i = \{c_{i,k}\}_{k=1}^K$,
but only one codeword \(c_i^*\) is dominant in the ML decision rule due to condition~\eqref{eq:cw_dominance} with all other codewords having negligible likelihood.

By Assumption 2, the network maps the input to a soft binary feature vector \(y \in [0,1]^n\), and the Hamming distance to each codeword \(c_{i,k}\) can be written as an affine function of \(y\). Hence, the logit associated with codeword \(c_{i,k}\) has the form
$$
z_{i,k}(y)
= \langle w_{i,k}, y \rangle + b_{i,k} + \beta(y).
$$

By Assumption 3, the total likelihood for class \(i\) is (up to normalization)
$$
p(y \mid \mathcal{C}_i)
\propto \sum_{k=1}^K \exp(z_{i,k}(y)).
$$

However, the condition~\eqref{eq:cw_dominance} implies by Lemma~\ref{lm:cw_dominance} that one codeword \(c_i^*\) overwhelmingly dominates the likelihood:
$$
\exp(z_{i,k}(y)) \ll \exp(z_{i,k^*}(y))
\qquad\text{for all } k\ne k^*,
$$
so that
$$
\sum_{k=1}^K \exp(z_{i,k}(y))
\simeq
\exp(z_{i,k^*}(y)).
$$

Since $\beta(y)$ is independent of $i$, $\mathrm{softmax}(z_i(y)+\beta(y))=\mathrm{softmax}(z_i(y))$.
Thus, the entire group of \(K\) neurons corresponding to class \(i\) can be replaced by a single effective neuron with parameters
$$
w_i^{\mathrm{eff}} := w_{i,k^*},
\qquad
b_i^{\mathrm{eff}} := b_{i,k^*},
$$
producing the logit
$$
z_i(y)
= \langle w_i^{\mathrm{eff}}, y\rangle + b_i^{\mathrm{eff}}.
$$

Thus, class-wise aggregation reduces the decoder to a standard linear layer producing logits $z_i(y)$.
\end{proof}

\begin{lemma}[Softmax preserves argmax]
\label{lm:ml_rule}
Let $z = (z_1, \ldots, z_M) \in \mathbb{R}^M$, and define the softmax probabilities
\begin{equation}
\label{sigma-z}
  \sigma(z_i)
  = \frac{\exp(z_i)}
         {\sum\limits_{j=1}^{M}\exp(z_j)}, \quad i=1,\ldots,M.
\end{equation}
Then
$$
\arg\max_i \sigma(z_i) = \arg\max_i z_i,
$$
i.~e., the set of indices that maximize the softmax components is identical to the set of indices that maximize the logits.
\end{lemma}
\begin{proof}
    The denominator $\sum_{j=1}^{M}\exp(z_j)$ is positive and does not depend on the index $i$; hence, it does not affect the ordering over $i$. Therefore
$$
\arg\max_i \sigma(z_i) = \arg\max_i \exp(z_i).
$$

The exponential function $\exp(\cdot)$ is strictly increasing on $\mathbb{R}$, so it preserves order:
$$
\arg\max_i \exp(z_i) = \arg\max_i z_i.
$$
Combining the two equations gives the claimed result.

\end{proof}

\begin{theorem}[Softmax Posterior as a Gibbs Distribution]\label{th:softmax_ml_decoder}
Assume that the conditions of Theorem~\ref{th:optimal_linear_sum} hold, so that each class admits an effective
(dominant) codeword $c_i$ for the given $y$.
Then
\begin{enumerate}
\item
Softmax probabilities take the Gibbs form with Hamming energy
\[
  p(i\mid y)
  = \frac{\exp\bigl(-\lambda\, d_H(y,c_i)\bigr)}
         {\sum\limits_{j=1}^{M}\exp\bigl(-\lambda\, d_H(y,c_j)\bigr)}.
\]

That is, the classifier computes a Gibbs distribution over classes using the Hamming distance as the energy function.

\item
The predicted class coincides with the ML decoder
$$
\hat{i}(y)=
\arg\max_{i} p(i \mid y) =
\arg\min_{i} d_{H}(y,c_i) =
\arg\max_{i} p(y \mid c_i),
$$
under the uniform class prior.

Thus, the softmax classifier implements maximum-likelihood decoding in the Hamming metric.
\end{enumerate}
\end{theorem}

\begin{proof}
Substituting the parametrization~\eqref{z_i-lb} into the softmax definition~\eqref{sigma-z},
\[
  p(i\mid y)
  = \frac{\exp\bigl(-\lambda\, d_H(y,c_i)+\beta(y)\bigr)}
         {\sum\limits_{j=1}^{M}\exp\bigl(-\lambda\, d_H(y,c_j)+\beta(y)\bigr)}
\]
Since \(\beta(y)\) is independent of \(i\), it cancels in the numerator and denominator:
\[
 p(i\mid y) = \frac{\exp\bigl(-\lambda\, d_H(y,c_i)\bigr)}
         {\sum\limits_{j=1}^{M}\exp\bigl(-\lambda\, d_H(y,c_j)\bigr)}.
\]
This establishes the Gibbs form and proves the first statement.

For the second statement, observe that the softmax preserves \rev{the} ordering of logits (Lemma~\ref{lm:ml_rule}):
$$
\arg\max_{i} p(i\mid y) =
\arg\max_{i} z_i(y).
$$
Using the parametrization~\eqref{z_i-lb},
$$
\arg\max_{i} z_i(y) =
\arg\max_{i} \big(-\lambda d_H(y,c_i)\big) =
\arg\min_{i} d_H(y,c_i).
$$
Under a uniform class prior, the ML decoder selects the class that maximizes the likelihood \(p(y\mid c_i)\), which, by Assumption 3, is proportional to \(\exp(-\lambda d_H(y,c_i))\). Thus the decision rule
$$
\hat{i}(y) = \arg\max_{i} p(i\mid y)
$$
is exactly the maximum-likelihood decoding rule.
\end{proof}

For an input \(y\) belonging to the true class \(t\), define the Hamming distances
$$
d_j := d_H(y, c_j), \quad
d_t := d_H(y, c_t),
$$
the \textit{relative Hamming distance gaps}
$$
\Delta_j^{(t)}(y) := d_j - d_t,
$$
and the \textit{Hamming margin} of the true class
\begin{equation}\label{eq:hamming-margin}
\gamma_t(y)
:= \min_{j \ne t} \Delta_j^{(t)}(y)
= \min_{j \ne t} \big( d_j - d_t \big).
\end{equation}

We assume that the ML decoder selects the correct class \(t\), i.~e., \(d_t \le d_j\) for all \(j\ne t\).

\begin{theorem}
\label{th:ce-hamming-margin}
If the conditions of Theorem~\ref{th:optimal_linear_sum} hold and the ML decoder assigns \(y\) to class \(t\),
then for sufficiently large positive Hamming margin \(\gamma_t(y)\)~\eqref{eq:hamming-margin}, the cross-entropy loss admits
the asymptotic expansion
$$
  \mathcal{L}(y,t)\;\approx\;(M-1)\,\exp(-\lambda \gamma_t(y)),
$$
up to terms of order $\exp(-2\lambda \gamma_t(y))$.
\end{theorem}

\begin{proof}
From Theorem~\ref{th:softmax_ml_decoder}, the class posterior is
\[
  p(i\mid y)
  = \frac{\exp(-\lambda d_i)}
         {\sum\limits_{j=1}^{M}\exp(-\lambda d_j)},
\]
hence,
\[
  \mathcal{L}(y,t)
  = -\ln p(t\mid y)
  = \ln\Bigl(1 + \sum_{j\neq t}\exp\bigl(-\lambda\Delta_j^t(y)\bigr)\Bigr).
\]

Because the ML decoder chooses $t$,
we have
$$
d_t \le d_j \quad \Rightarrow \quad
\Delta_j^{(t)}(y) \ge 0, \qquad \gamma_t(y) \ge 0.
$$

Furthermore, for each \(j\ne t\),
$$
\gamma_t(y) \le \Delta_j^{(t)}(y)
\quad\Rightarrow\quad
\exp({-\lambda \Delta_j^{(t)}(y)})
\le
\exp({-\lambda \gamma_t(y)}).
$$

Thus the sum satisfies the bounds
$$
\exp({-\lambda \gamma_t(y)})
\;\le\;
\sum\limits_{j\ne t} \exp({-\lambda \Delta_j^{(t)}(y)})
\;\le\;
(M-1)\,\exp({-\lambda \gamma_t(y)}).
$$

Define
$$
S(y)
:=
\sum_{j\ne t} \exp({-\lambda \Delta_j^{(t)}(y)}),
$$
so
$$
S(y)
\in
\big[\exp({-\lambda \gamma_t(y)}), (M-1)\exp(-\lambda \gamma_t(y))\big],
$$
and therefore
\[
  \mathcal{L}(y,t) = \ln\bigl(1+S(y)\bigr).
\]

As $\gamma_t(y) \to \infty$, we have $S(y)\to 0$; hence, the Taylor expansion of the function \(\ln(1+S)\) gives
\[
  \ln\bigl(1+S\bigr)
  = S + O\bigl(S^2\bigr).
\]
Thus,
\[
  \mathcal{L}(y,t)
  = S(y) + O\bigl(S^2(y)\bigr),
\]
and using the bounds on \(S(y)\),
\[
  \mathcal{L}(y,t)
  = (M-1)\exp({-\lambda\gamma_t(y)}) + O\bigl(\exp({-2\lambda\gamma_t(y)})\bigr).
\]
This completes the proof.
\end{proof}

Theorem~\ref{th:optimal_linear_sum} shows that the group-wise aggregation of
codeword-related neurons can always be absorbed into the weights of a single
linear layer with \(M\) outputs. Despite the presence of \(K\) codewords per
class, under the likelihood model~\eqref{g||}, a single dominant codeword
effectively determines the decision for a given observation, so the decoder
operates as a dynamic choice of one codeword from a fixed codebook.

Theorem~\ref{th:softmax_ml_decoder} makes explicit that the standard training
objective, the cross-entropy at the output of a softmax layer, is equivalent
to minimizing a smooth surrogate of the Hamming distance between the feature
vector \(y\) and the dominant codeword \(c_t\) of the true class. This
connects the continuous optimization of neural networks with the discrete
geometry of optimal FEC codes in feature space.
 
\section{Experimental evaluation}
\label{sec:experiments}

This section presents empirical evidence supporting the discrete, channel-based interpretation of neural classifiers. As noted previously, prior Neural Collapse studies established, \rev{for cross-entropy}~\cite{papyan2020neuralcollapse} \rev{and MSE}~\cite{Han-2021-NC-MSE} \rev{losses,} that last-layer features and classifier weights converge to a simplex-like Euclidean geometry under standard training. Taking this geometry as a foundational premise, the analysis investigates complementary questions that remain unaddressed in the Neural Collapse literature:

\begin{enumerate}
  \item Does standard cross-entropy training recover a maximum-likelihood Hamming
        decoder on a purely synthetic Bernoulli channel with a fixed binary codebook?
  \item Are \rev{the} pre-fully connected layer (pre-FC) outputs \rev{of} a sufficiently large network (regnet\_x\_3\_2gf) \rev{informative enough} 
  to support approximate inversion to recognizable images when matched via the symmetric Kullback--Leibler loss $\mathcal{L}_{\mathrm{KL}}$ defined in Eq.~\ref{eq:inversion-loss}, as predicted 
  by the proposed channel model?
\end{enumerate}

All experiments demonstrate network behavior consistent with the predictions of Theorems~\ref{th:softmax_ml_decoder}--\ref{th:ce-hamming-margin}. Specifically, logits function as smooth surrogates for the Hamming distance, while the learned codebooks display genuine error-correcting capabilities in the presence of synthetic bit-flip noise.

\subsection{\texorpdfstring{Synthetic Bernoulli code with learnable linear decoder}{Synthetic Bernoulli code with learnable linear decoder}}
\label{subsec:exp-synth-results}
The first scenario, devoted to studying learnable linear decoders for synthetic Bernoulli codebooks, addresses a fully synthetic setting in which inputs are generated by transmitting class-dependent binary codewords $c_i \in \{0, 1\}^n$ over a binary symmetric channel (BSC), and only a linear softmax layer is trained. The experiment uses $M=32$ classes, $K \in \{1, 8, 16\}$ codewords per class, code length $n=256$, target minimum-distance threshold $d_{min}/n \ge 0.3$, and BSC crossover probability $q \in \{0.1, 0.2, 0.3, 0.4\}$.

The value $n=256$ was chosen primarily to make constrained random-codebook generation feasible: it is large enough for Hamming distances between random binary vectors to concentrate, while still small enough for rejection/greedy construction under a prescribed minimum-distance constraint. This scale is also close to classical length-255 block codes used in communications, for example the CCSDS Reed--Solomon $(255,223)$ space-telemetry code~\cite{Costello-1992-CCSDS}. The values $M=32$ and the threshold $d_{min}/n\ge0.3$ were chosen jointly: the logged realized minima are $d_{min}=107$ for $K=1$ (codebook size $32$) and $d_{min}=94$ for $K\in\{8,16\}$ (codebook sizes $256$ and $512$), while the average pairwise distance remains near $128\approx0.5n$. As a mild reference point, an optimal binary linear $[256,5,131]$ code gives $32$ codewords with $d_{min}=131\approx0.512n$~\cite{Grassl-CodeTables-256-5}; hence the generated random codebooks are well separated but not near-optimal in the worst-case-distance sense.

\begin{figure}[t]
  \centering
  \begin{tikzpicture}
\begin{axis}[
    width=0.95\linewidth,
    height=0.55\linewidth,
    xlabel={BSC flip $q$},
    ylabel={Accuracy},
    ymin=0, ymax=1.05,
    ymajorgrids=true,
    xmajorgrids=true,
    legend style={at={(0.02,0.02)},anchor=south west,draw=none,fill=none},
    mark size=1.8pt,
]
\addplot+[ml,grpOne,mymarks] coordinates {(0.1,1.00) (0.2,1.00) (0.3,1.00) (0.4,0.85)};
\addlegendentry{ML K=1}
\addplot+[nn,grpOne,mymarks] coordinates {(0.1,1.00) (0.2,1.00) (0.3,1.00) (0.4,0.83)};
\addlegendentry{NN K=1}

\addplot+[ml,grpEight,mymarks] coordinates {(0.1,1.00) (0.2,1.00) (0.3,1.00) (0.4,0.66)};
\addlegendentry{ML K=8}
\addplot+[nn,grpEight,mymarks] coordinates {(0.1,1.00) (0.2,0.97) (0.3,0.70) (0.4,0.24)};
\addlegendentry{NN K=8}

\addplot+[ml,grpSixteen,mymarks] coordinates {(0.1,1.00) (0.2,1.00) (0.3,1.00) (0.4,0.58)};
\addlegendentry{ML K=16}
\addplot+[nn,grpSixteen,mymarks] coordinates {(0.1,0.99) (0.2,0.79) (0.3,0.43) (0.4,0.15)};
\addlegendentry{NN K=16}

\end{axis}
\end{tikzpicture}   \caption{Synthetic classification accuracy as a function of noise level \(q\) for ML and learned softmax decoders.}
  \label{fig:synth-acc-curve}
\end{figure}
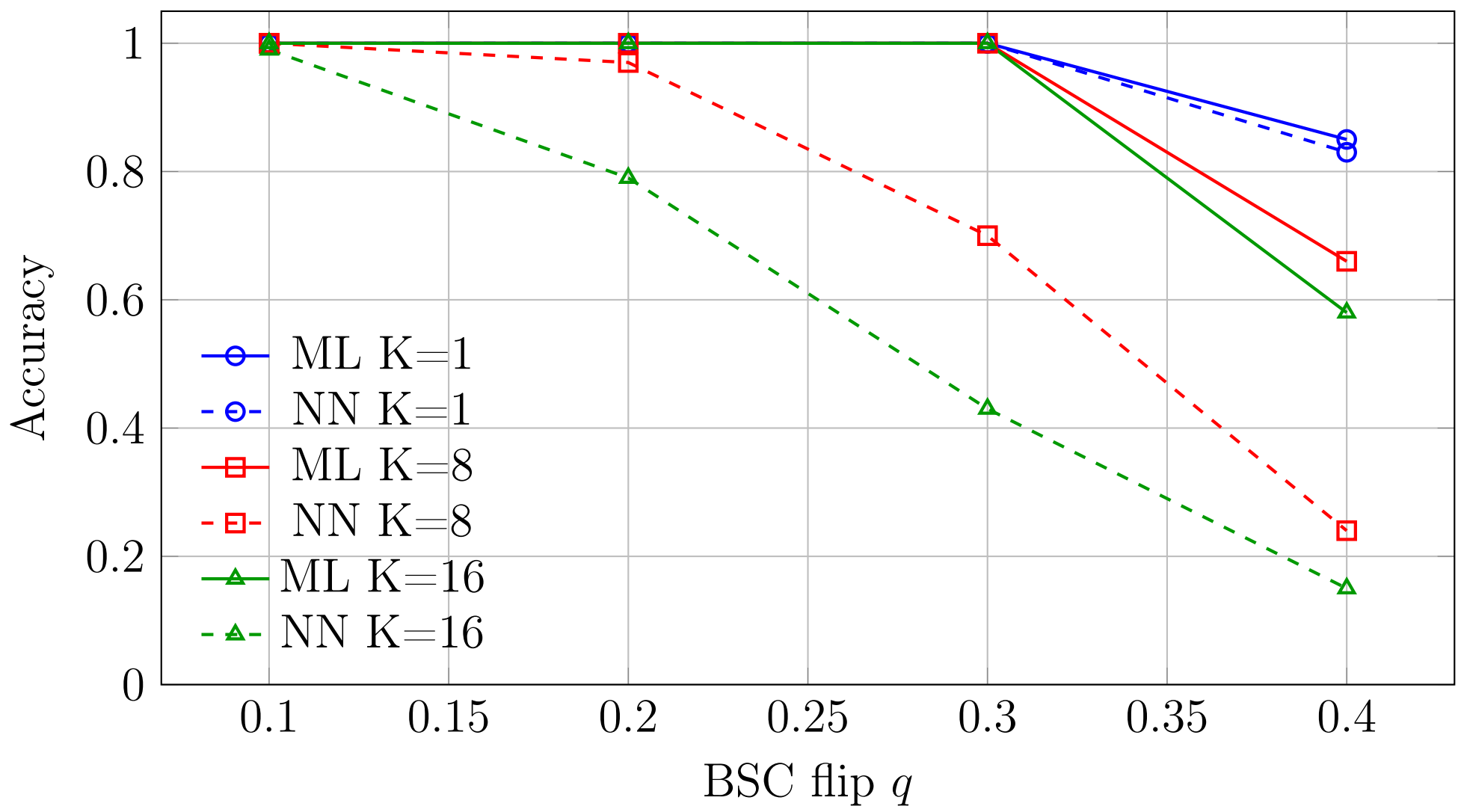

The \rev{crossover-probability set was selected relative to the code rate.} The class-label rate is $R_{\mathrm{cls}}=\log_2 M/n=5/256\approx0.0195$. Decoding the full union codebook corresponds to rates $R_{\mathrm{code}}=\log_2(MK)/n \in \{0.0195, 0.0313, 0.0352\}$ for $K\in\{1, 8, 16\}$. For a BSC, the Shannon capacity is $C(q)=1-H_2(q)$. For the tested set $q\in\{0.1,0.2,0.3,0.4\}$, this gives
\[
  C(q)\in\{0.531, 0.278, 0.119, 0.029\}.
\]
These asymptotic values overstate what a length-$n=256$ code achieves. Correcting $C(q)$ by the BSC channel dispersion $V(q)=q(1-q)\log_2^2\tfrac{1-q}{q}$ gives the finite-length capacity~\cite{Polyanskiy-2010-FiniteBlocklength}
\[
  C_n(q)=C(q)-\sqrt{V(q)/n}\,Q^{-1}(\varepsilon)+\tfrac{\log_2 n}{2n},
\]
which for a target error $\varepsilon=10^{-2}$ evaluates to
\[
  C_n(q)\in\{0.408,\,0.177,\,0.053,\,0.003\}.
\]
For $q\in\{0.1,0.2,0.3\}$ this exceeds every codebook rate, and the ML decoder reaches near-unit accuracy. At $q=0.4$ it collapses to $C_n=0.003$, below even the class-label rate $R_{\mathrm{cls}}=0.0195$: the channel cannot support reliable decoding at this block length, and the maximum-likelihood decoder saturates near $0.8$ accuracy already at $K=1$. Training a single linear layer via cross-entropy on $10^6$ synthetic samples yields the following observations:

\paragraph{\textbf{Decoder agreement}}
On an independent test set of $10^5$ samples, the learned softmax decoder
$\hat{i}_{\mathrm{soft}}(y) = \arg\max_i z_i(y)$ and the Bayes-optimal Hamming
decoder $\hat{i}_{\mathrm{ML}}(y) = \arg\min_i d_H(y,c_i)$ demonstrate 
similar performance at \(K=1\). With a higher number of codewords per class, the softmax decoder degrades much faster with increasing noise level, as shown in Fig.~\ref{fig:synth-acc-curve}.
This confirms that standard cross-entropy training of a linear layer essentially recovers a
maximum-likelihood Hamming decoder for the underlying binary channel, for moderate values of \(q\) and \(K\). 

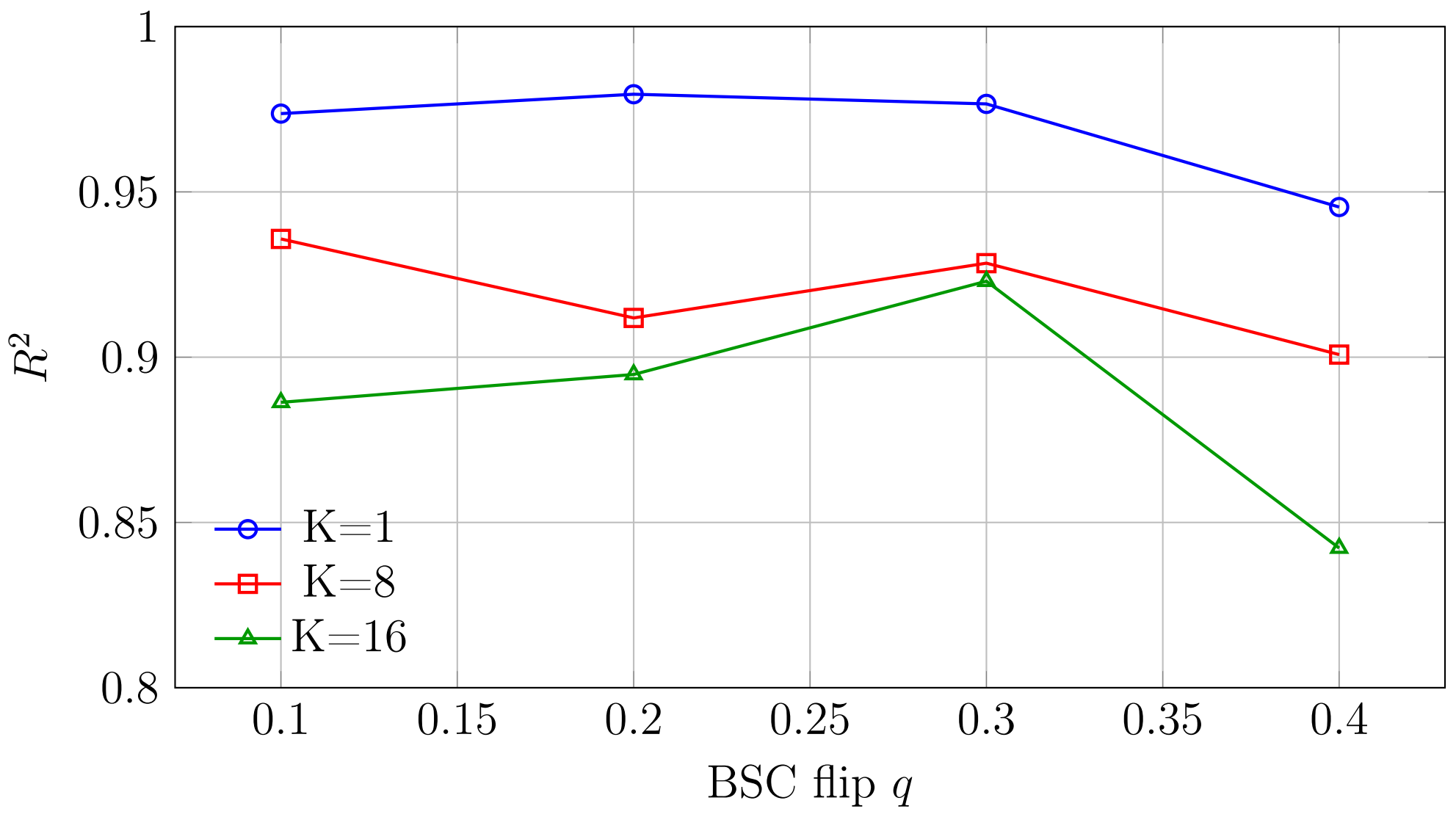
\begin{figure}[t]
  \centering
  \begin{tikzpicture}
\begin{axis}[
    width=0.95\linewidth,
    height=0.55\linewidth,
    xlabel={BSC flip $q$},
    ylabel={$R^2$},
    ymin=0.8, ymax=1.0,
    ymajorgrids=true,
    xmajorgrids=true,
    legend style={at={(0.02,0.02)},anchor=south west,draw=none,fill=none},
]

\addplot+[ml,grpOne,mymarks] coordinates {
  (0.1,0.973686)
  (0.2,0.979547)
  (0.3,0.976621)
  (0.4,0.945421)
};
\addlegendentry{K=1}

\addplot+[ml,grpEight,mymarks] coordinates {
  (0.1,0.935772)
  (0.2,0.911871)
  (0.3,0.928439)
  (0.4,0.900784)
};
\addlegendentry{K=8}

\addplot+[ml,grpSixteen,mymarks] coordinates {
  (0.1,0.886338)
  (0.2,0.894763)
  (0.3,0.922987)
  (0.4,0.842260)
};
\addlegendentry{K=16}

\end{axis}
\end{tikzpicture}   \caption{Synthetic classification \(R^2\) of a linear fit of logits to Hamming distance for learned softmax decoders.}
  \label{fig:synth-logits-hamming}
\end{figure}

\paragraph{\textbf{Logits vs. Hamming distance}}
For each pair of classes $(i,j)$ and each test sample $y$, \rev{define}
\[
  \Delta z_{i,j}(y) = z_j(y) - z_i(y), \qquad
\Delta d_{i,j}(y) = \sum_{k=1}^K\Big(d_H(y,c_{j,k}) - d_H(y,c_{i,k})\Big).
\]
Pooling all pairs and samples together, a global linear fit
$\Delta z \approx -\lambda\, \Delta d + \beta$ achieves high values of the coefficient of determination ($R^2$), as shown in Fig.~\ref{fig:synth-logits-hamming}, 
and empirically supports the affine form of the decoder in Hamming distance
derived in Theorem~\ref{th:softmax_ml_decoder}.

Empirical observations in Figures~\ref{fig:synth-acc-curve} and~\ref{fig:synth-logits-hamming} indicate that decoder behavior aligns with Assumption 3 from Section~\ref{sec:softmax} under conditions of low BSC noise ($q$) and small codebook multiplicity ($K$). However, for larger values of $q$ and $K$, this assumption fails to hold, suggesting that multiple codewords can contribute significantly to the class likelihood.
The synthetic Bernoulli experiment therefore verifies the idealized assumptions in a setting where the channel is exactly controlled, whereas real neural networks should be interpreted through local approximations of the same assumptions. Regularization, feature decoupling across deep layers, and representation learning tend to suppress nuisance directions and concentrate class-relevant variation, making the residual feature noise closer to isotropic and often leaving a small number of codewords dominant for the final decoder. This view is empirically supported by Neural Collapse, where late-stage training concentrates same-class features around class means and aligns classifier weights with \rev{these means}~\cite{papyan2020neuralcollapse}.
 \subsection{\texorpdfstring{Inverting pre-FC codewords in a pretrained regnet\_x\_3\_2gf}{Inverting pre-FC codewords in a pretrained regnet x 3 2gf}}
\label{subsec:exp-regnet-results}
The following experiments are conducted by inverting trunk-output (pre-FC) codewords in a pretrained Torchvision regnet\_x\_3\_2gf network.
\paragraph{\textbf{Benchmarking Against Prior Inversion Strategies}}
Prior works on neural model inversion are typically based either on Euclidean feature regression~\cite{MahendranVedaldi2015Inverting} or on restricting reconstructions to the range of a pretrained GAN while optimizing the latent code, \rev{either for class-conditional synthesis}~\cite{Nguyen2017PPGN} \rev{or for model-inversion attacks}~\cite{Chen2021KEDMI}. 
Both strategies rely on strong inductive biases to prioritize realistic images, achieved either through explicit priors or regularized loss functions. Consequently, for inputs that match the training data, these methods can achieve near-perfect instance-level reconstruction.
Moreover, a purely pixel-space, GAN-free inversion based on Euclidean feature matching is generally viewed as unstable or ineffective for sufficiently deep networks without additional priors.

In contrast, the proposed reconstruction in this paper is driven by an information-geometric objective: the symmetric \(\mathcal{L}_{\mathrm{KL}}\) loss between softmax-normalized features. This criterion considers probability vectors on the simplex rather than feature coordinates in \(\mathbb{R}^n\) and, therefore, is not intended for pixel-level recovery of a particular input. Empirically, the same objective remains effective on deep architectures without introducing an external generative model.
Theorem~\ref{th:softmax_ml_decoder} frames the proposed reconstruction as an activation-maximization procedure analogous to DeepDream~\cite{Mordvintsev2015InceptionismDeepDream}. However, unlike standard approaches that target a hand-picked unit, the ``activation'' here is defined by the class posterior (or its pre-FC softmax evidence).

The generative sufficiency of pre-FC codewords is evaluated using the pretrained \rev{Torchvision}~\cite{marcel2010torchvision} regnet\_x\_3\_2gf \rev{model}~\cite{Radosavovic2020DesigningRegNet} on ImageNet~\cite{Russakovsky2015ILSVRC}.
Target features are extracted from the frozen ReLU model. During inversion, the same pretrained weights are used with ReLU activations replaced by Softplus to provide a smooth optimization path; classification checks use the original frozen ReLU model. An image \(x\) is reconstructed by optimizing pixels so that the induced
feature distribution \(p_{\theta}(x)=\mathrm{softmax}(F(x))\) matches the target distribution \(p_{\theta}(x^*)\)
under \(\mathcal{L}_{\mathrm{KL}}\), with regularization using only the total variation $TV(x)=\|\nabla x\|_{1}$,
which serves as a physically motivated plausibility constraint that suppresses non-physical high-frequency oscillations while preserving
edges~\cite{Rudin1992ROF}. Authentic images are effectively modeled as functions of bounded variation.
Due to finite optical resolution and sensor integration, the images are predominantly piecewise smooth with sparse edges, resulting
in spatial gradients concentrated on a small subset of pixels.

Pre-FC features are inverted on the ImageNet validation set (unseen during training) using resize-to-256 and 224-by-224 center-crop preprocessing. For numerical stability in the inversion model, ReLU activations are replaced with Softplus ($\beta=3$).

The optimization objective consists of the following components:
\begin{itemize}
    \item [(i)] The \(\mathcal{L}_{\mathrm{KL}}\) term defined in Eq.~\ref{eq:inversion-loss}.
\item [(ii)] A feature-centering constraint implemented as an MSE penalty on the feature means.
\item [(iii)] An $l_1$ total-variation prior, lower-bounded by the total variation of the original image.

\end{itemize}

Let \(F(x)\in\mathbb{R}^{n}\), where $n$ is the number of pre-FC layer output channels. The loss notation used below is defined by
\begin{equation}
\label{eq:inversion-loss}
\begin{aligned}
\mathcal{L}_{\mathrm{KL}}(x,x^*)
&= \tfrac12\big[D_\mathrm{KL}(p_{\theta}(x)\,\|\,p_{\theta}(x^*))
+ D_\mathrm{KL}(p_{\theta}(x^*)\,\|\,p_{\theta}(x))\big]\\
&= \tfrac12D_J\!\left(p_{\theta}(x),p_{\theta}(x^*)\right),\\
\mathcal{L}_{\mathrm{MSE}} &= \big\|\mathbb{E} F(x) - \mathbb{E} F(x^*)\big\|_2^2,\\
\mathcal{L}_{TV} &= \max\!\big(TV(x) - 0.4\,TV(x^*),\, 0\big),\\
\mathcal{L}(x) &= \mathcal{L}_{\mathrm{KL}}(x,x^*)+ \lambda_2\,\mathcal{L}_{\mathrm{MSE}}+ \lambda_{TV}\,\mathcal{L}_{TV}.
\end{aligned}
\end{equation}

Optimization is performed for 4000 steps with Adam and cosine learning-rate annealing from an initial learning rate of $0.5$, a total-variation weight $\lambda_{TV}=2.5\times 10^{-3}$, a feature-mean MSE weight $\lambda_2=10$, and Gaussian initialization
\(x\sim\mathcal{N}(0,\,0.01^2)\). After each iteration, the optimized image \(x\) is re-standardized to zero mean and unit standard deviation, i.~e., \((\mathbb{E}[x],\operatorname{std}(x))=(0,1)\).

\begin{figure}[t]
  \centering
  \includegraphics[width=\linewidth]{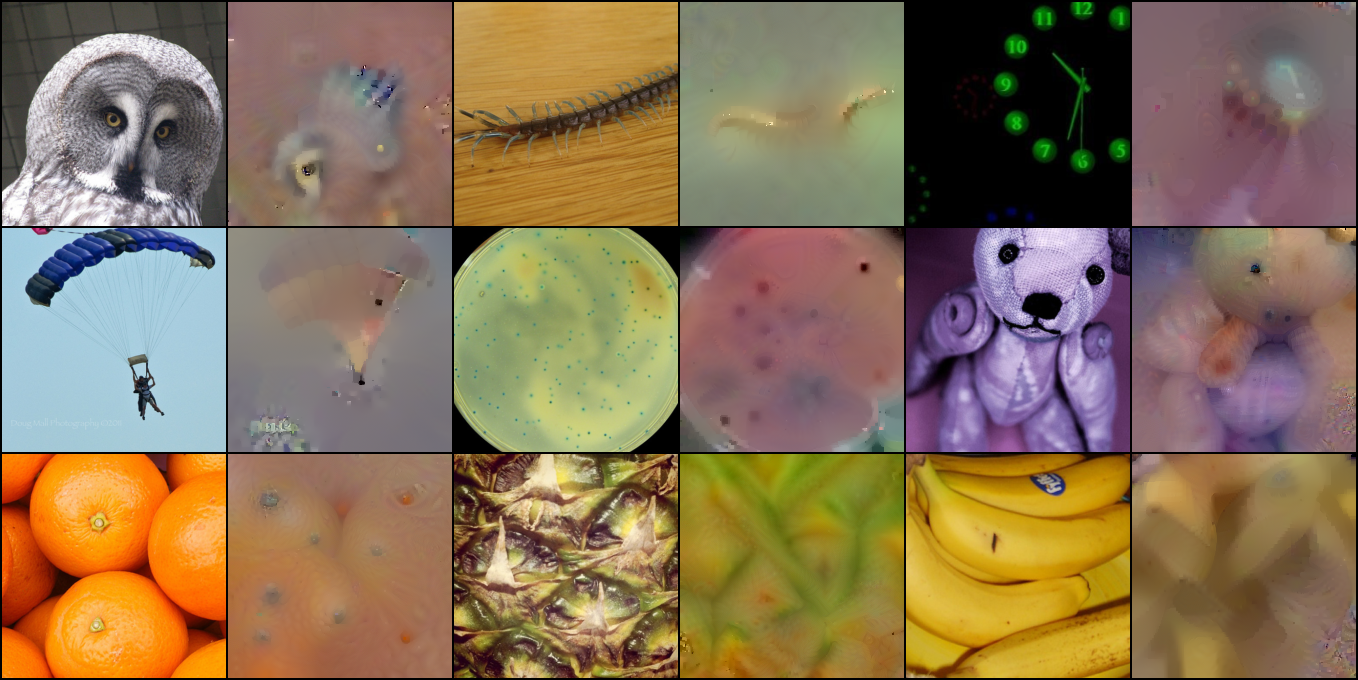}
  \caption{Results obtained by inverting pre-FC codewords in a pretrained regnet\_x\_3\_2gf network.}
  \label{fig:regnet-recon}
\end{figure}

For the inversion experiment, we handpicked classes 24, 79, 409, 701, 712, 850, 950, 953, and 954. Candidate validation images are first filtered to samples correctly classified by the frozen model, ranked by the top-1 minus top-2 logit gap, balanced across the selected classes, and then inverted. The displayed examples are selected by the smallest final \(\mathcal{L}_{\mathrm{KL}}\) loss. A sample of the reconstruction results for these classes is shown in Fig.~\ref{fig:regnet-recon}.

The TV prior proves to be essential; in its absence, the optimization degenerates into high-frequency noise-like artifacts. Moreover, the \(\mathcal{L}_{\mathrm{KL}}\) term alone is under-constrained, as softmax-based divergences are invariant to additive shifts in the logits. Introducing the feature-mean constraint resolves this degeneracy, thereby restoring reliable convergence.

While the resulting reconstructions are not pixel-identical, they exhibit high semantic fidelity. Global color cues are accurately preserved; for instance, oranges retain their orange hue and bananas their yellow-green tint. Collectively, these findings validate the hypothesis that the pre-FC representation partitions the input space into smooth, semantically coherent clusters that map to proximate codewords in feature space.

Fig.~\ref{fig:regnet-recon}
presents inversion results for regnet\_x\_3\_2gf, providing an operational analog of Plato's distinction between appearance and form. The figure compares original ImageNet validation images with reconstructed images obtained by minimizing \(\mathcal{L}_{\mathrm{KL}}\) between pre-FC feature distributions, augmented with a total-variation prior. In this setting, many distinct pixel-level realizations $x$~(appearances) correspond to essentially the same network-defined internal representation $y$~(form). Reconstructions performed via \(\mathcal{L}_{\mathrm{KL}}\) matching in representation space, while constrained to physically plausible natural images through the TV prior, yield diverse images that preserve class-level semantics without reproducing exact pixel configurations. This demonstrates that the learned representation effectively operates as a semantic equivalence class.

\paragraph{\textbf{Objective CLIP evaluation}}
\label{subsec:clip-objective-evaluation}

CLIP is a vision-language representation model~\cite{Radford2021CLIP} whose image encoder maps semantically related images to nearby embedding vectors, making it suitable for measuring whether a reconstruction preserves the source image semantics. For each model out of the 15 tested, the full ImageNet validation set is used as the candidate pool; a shuffled subset of \(5000\) images is screened by the frozen classifier, only correctly classified images are retained, and these candidates are sorted by the top-1 minus top-2 logit margin. The first \(48\) high-margin candidates form the inversion batch. The evaluator encodes both originals and reconstructions using OpenCLIP ViT-B-32 with OpenAI-pretrained weights, \(L_2\)-normalizes the image embeddings, and forms the image-image similarity matrix \(S_{ij}=\langle e_i^{\mathrm{orig}},e_j^{\mathrm{rec}}\rangle\).
To reduce row and column bias, the reported metrics use the row/column-normalized matrix \(S_{\mathrm{norm}}\) from the experiment logs. Recall@ \(k\) measures whether the paired reconstruction is among the top \(k\) retrieved reconstructions for the corresponding original. MRR (mean reciprocal rank) averages \(1/r_i\) over the \(48\) originals, where \(r_i\) is the rank of the paired reconstruction for original \(i\); larger values mean that paired reconstructions are retrieved earlier. The matched and nonmatched columns report the diagonal and off-diagonal means of \(S_{\mathrm{norm}}\), and the \(p\)-value is a one-sided Wilcoxon signed-rank test applied to the per-image differences between the matched score and the same-row nonmatching mean. Its null hypothesis is that matched reconstructions are not systematically more similar to their originals than mismatched reconstructions; a small \(p\)-value therefore indicates statistically supported semantic pairing, while the Gap, R@1, R@5, and MRR columns quantify the effect size.
The $\mathcal{L}_{\mathrm{KL},0}$ and $\mathcal{L}_{\mathrm{KL},T}$ columns report the initial and final KL loss values of Eq.~\ref{eq:inversion-loss}.
Inv. Acc@1 reports the top-1 accuracy obtained by re-classifying the \(48\) reconstructed images with the same frozen model and comparing predictions with the original ImageNet labels. The Acc@1 column reports the official Torchvision ImageNet-1K single-crop top-1 accuracy for the corresponding default pretrained weights~\cite{torchvision-models-weights}.

\begin{table}[t]
\centering
\tiny
\setlength{\tabcolsep}{1.0pt}
\resizebox{\columnwidth}{!}{\begin{tabular}{l|rr|rr|rr|l|rr|rr}
\hline
\textbf{Model} & \textbf{R@1} & \textbf{R@5} & \textbf{Match} & \shortstack{\textbf{Non}\\\textbf{match}} & \textbf{Gap} & \textbf{MRR} & \textbf{\(p\)} & \textbf{\(\mathcal{L}_{\mathrm{KL},0}\)} & \textbf{\(\mathcal{L}_{\mathrm{KL},T}\)} & \shortstack{\textbf{Inv.}\\\textbf{Acc@1}} & \textbf{Acc@1} \\
\hline
regnet\_x\_400mf & 33.3 & 72.9 & 0.662 & 0.605 & 0.056 & 0.504 & $10^{-13}$ & 224.8 & 7.7 & 95.8 & 74.9 \\
regnet\_x\_800mf & 33.3 & 52.1 & 0.678 & 0.623 & 0.055 & 0.450 & $10^{-10}$ & 257.0 & 26.5 & 91.7 & 77.5 \\
regnet\_x\_1\_6gf & 27.1 & 75.0 & 0.660 & 0.612 & 0.048 & 0.448 & $10^{-13}$ & 243.3 & 22.9 & 97.9 & 79.7 \\
regnet\_x\_3\_2gf & 43.8 & 56.2 & 0.658 & 0.603 & 0.056 & 0.515 & $10^{-11}$ & 234.9 & 16.0 & 100.0 & 81.2 \\
regnet\_x\_8gf & 33.3 & 68.8 & 0.639 & 0.597 & 0.042 & 0.491 & $10^{-11}$ & 224.2 & 56.1 & 83.3 & 81.7 \\
\hline
regnet\_y\_400mf & 31.2 & 56.2 & 0.672 & 0.613 & 0.059 & 0.451 & $10^{-8}$ & 262.6 & 12.1 & 93.8 & 75.8 \\
regnet\_y\_800mf & 31.2 & 66.7 & 0.626 & 0.587 & 0.039 & 0.449 & $10^{-11}$ & 343.0 & 14.6 & 100.0 & 78.8 \\
regnet\_y\_1\_6gf & 31.2 & 47.9 & 0.639 & 0.607 & 0.032 & 0.410 & $10^{-8}$ & 244.0 & 17.4 & 97.9 & 80.9 \\
regnet\_y\_3\_2gf & 22.9 & 58.3 & 0.612 & 0.576 & 0.035 & 0.384 & $10^{-12}$ & 230.0 & 12.8 & 97.9 & 82.0 \\
regnet\_y\_8gf & 8.3 & 50.0 & 0.612 & 0.584 & 0.027 & 0.271 & $10^{-12}$ & 193.3 & 27.1 & 89.6 & 82.8 \\
\hline
resnet18 & 6.2 & 25.0 & 0.553 & 0.545 & 0.008 & 0.172 & $10^{-4}$ & 412.5 & 160.7 & 10.4 & 69.8 \\
resnet34 & 0.0 & 8.3 & 0.539 & 0.538 & 0.001 & 0.078 & $10^{-1}$ & 1001.8 & 870.4 & 0.0 & 73.3 \\
resnet50 & 29.2 & 58.3 & 0.640 & 0.597 & 0.043 & 0.419 & $10^{-9}$ & 227.0 & 36.7 & 93.8 & 80.9 \\
resnet101 & 14.6 & 39.6 & 0.618 & 0.589 & 0.029 & 0.278 & $10^{-6}$ & 250.7 & 79.6 & 75.0 & 81.9 \\
resnet152 & 20.8 & 45.8 & 0.624 & 0.588 & 0.036 & 0.349 & $10^{-8}$ & 218.2 & 56.4 & 85.4 & 82.3 \\
\hline
\end{tabular}}
\caption{Objective CLIP evaluation of all model-specific clip48 experiments. Acc@1, Inv. Acc@1, R@1, and R@5 are percentages. All CLIP metrics are computed on \(48\) original/reconstruction pairs using the normalized \(S_{\mathrm{norm}}\) block. The $\mathcal{L}_{\mathrm{KL}}$ columns report initial and final KL losses defined in Eq.~\ref{eq:inversion-loss}.}
\label{tab:clip48-objective}
\end{table}

The CLIP evaluation links the inversion experiment to both interpretations developed above. First, the optimization minimizes only an information-geometric distance between internal feature distributions, together with a weak TV prior, yet the retrieved reconstructions preserve semantic content in CLIP space. This supports the Plato-style view that the network representation stores a semantic form rather than a unique pixel-level appearance. Second, in FEC terms, the positive matched/nonmatched gaps indicate that the reconstructed samples remain inside a semantic decoding neighborhood of the original codewords. The \(p\)-values support this interpretation for almost all models: the very small values reject the null that matched reconstructions are no more similar than mismatches, whereas ResNet-34 \((p=0.3002)\) does not provide such evidence and also has nearly zero Gap and R@1. The same qualitative behavior appears across the three tested model families, RegNet-X, RegNet-Y, and ResNet: successful runs reduce the \(\mathcal{L}_{\mathrm{KL}}\) inversion loss and produce positive CLIP semantic gaps. The size dependence is also non-monotone. The best Recall@1 in the tested grid is obtained by regnet\_x\_3\_2gf for RegNet-X, by the compact-to-mid RegNet-Y models rather than the largest RegNet-Y model, and by ResNet-50 for ResNet. This suggests an empirical model-size sweet spot, where representation capacity is high enough to encode semantic structure but not so large that the inversion protocol becomes dominated by overly specific or less reusable feature detail. At this point, the reference ImageNet Acc@1 is also close to saturation within the corresponding family, so further gains in classifier accuracy do not necessarily improve semantic inversion retrieval.

\paragraph{\textbf{Additional Reconstructions}}
An extended set of original and reconstructed image pairs produced by the same inversion procedure is provided in Fig.~\ref{fig:regnet-recon-ext} in the Appendix for reference.
 
\section{Conclusion} 

This paper has shown that the classification problem in deep neural networks admits a natural and rigorous interpretation as the decoding of an error-correcting code in the sense of Shannon's information theory. Under this perspective, many standard components of modern neural architectures, such as softmax outputs, cross-entropy loss, and linear classifiers, can be interpreted as direct analogues of classical maximum-likelihood (ML) decoding rules.

A key result is that the commonly used loss functions in supervised classification admit a clear coding-theoretic meaning. For a binary memoryless channel, the Jeffreys divergence between the true soft labels and the output of the optimal softmax classifier is proportional to the Hamming distance between true and predicted binary features; in the symmetric case, the same holds for the one-sided Kullback-Leibler divergence. In practical networks, where the final classifier is implemented by a fully connected linear layer followed by softmax, the logits can be interpreted as linearly transformed (negative) Hamming distances to the dominant class codewords. Minimizing the cross-entropy loss, therefore, becomes equivalent to minimizing the distance to the correct codeword. This provides a principled explanation for the effectiveness of the softmax-cross-entropy framework: it is an approximate realization of the classical criterion of minimizing the number of decoding errors.

From this perspective, the proposed framework extends and refines the classical theory of Error-Correcting Output Codes (ECOC). It is known that traditional ECOC approaches enhance robustness by choosing an appropriate code matrix for the output layer, but they are constrained by the explicit choice of the codebook and by operating only at the classifier head. In contrast, the present formulation shows that the entire deep architecture implicitly implements a codebook, with the code's optimality emerging from the requirements of FEC decoding under isotropic noise in feature space. Thus, ECOC can be viewed as a special case within a broader FEC-based interpretation of neural classification.
This viewpoint also leads to a new interpretation of the Neural Collapse phenomenon. The simplex geometry of class means and classifier weights observed in Neural Collapse arises naturally from coding-theoretic considerations: good FEC codes in Euclidean space maximize pairwise distances between codewords under a fixed energy budget, leading to simplex-like configurations. Similarly, the isotropic Gaussian geometry favored in LeJEPA can be understood using the same intuition: an optimal latent space should be symmetric, isotropic, and well-separated to enable reliable decoding. In contrast to prior work \rev{on Neural Collapse}~\cite{papyan2020neuralcollapse} \rev{and LeJEPA}~\cite{Balestriero-2025-LeJEPA}, this geometry is not postulated but derived from fundamental requirements of robustness and optimal decoding.

The proposed framework has several practical implications. First, it helps explain the empirical success of neural quantization methods: if the architecture has sufficient coding-theoretic capacity, then weights and activations can be quantized without significant degradation in accuracy. Second, replacing cross-entropy with the Jeffreys divergence in knowledge distillation aligns the distillation objective more closely with the FEC criterion of minimizing bit-level errors, potentially improving the transfer of soft labels. Third, the FEC interpretation provides principled guidance for architectural design: increasing depth and width corresponds to increasing code length and redundancy, while regularization influences the geometry of the emergent codewords, encouraging simplex-like or isotropic configurations that improve robustness. Finally, the approach naturally extends to self-supervised encoder models such as LeJEPA and variational autoencoders. If the latent representation \(h_\theta(x)\) is understood as a codeword produced by an optimal encoder, then the downstream classifier \(p_\psi(y \mid h_\theta(x))\) performs ML decoding in the presence of noise and quantization.

The empirical results support a channel-based view of neural classifiers, in which learned representations act as robust, error-correcting codewords. Across both synthetic and real-world settings, cross-entropy training induces decoding behavior consistent with maximum-likelihood inference, while pre-FC representations define smooth, semantically coherent equivalence classes, indicating that robustness and generalization arise from operating on discrete semantic forms rather than exact pixel details. 

Consequently, the latent geometries derived in LeJEPA and the geometric structures observed in Neural Collapse correspond to different manifestations of the same underlying FEC principle: good representations are precisely those that behave as well-designed codewords enabling reliable decoding.
In analogy with Shannon's classical results, this work provides a conceptual benchmark for understanding when quantization can be performed without loss of accuracy. However, the problem of identifying architectural conditions that guarantee such quantization remains open. It is hoped that the FEC-based perspective introduced here will serve as a valuable foundation for both theoretical investigation and the practical design of quantized, energy-efficient, and noise-robust neural models.
 
\bibliographystyle{elsarticle-num}

\appendix
\section{Extended semantic comparison}

\begin{figure}[t]
  \centering
  \includegraphics[width=0.85\linewidth]{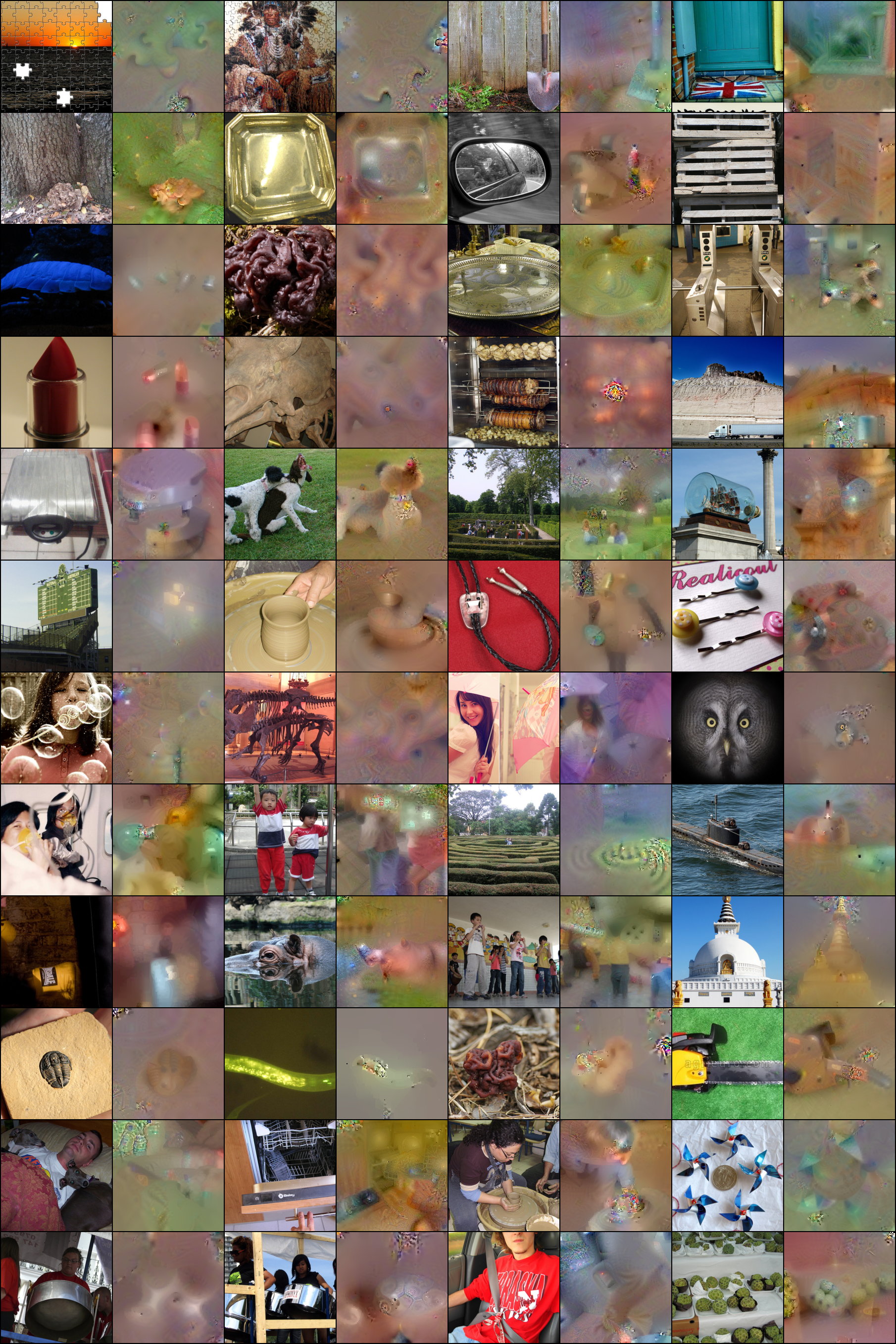}
  
  \caption{regnet\_x\_3\_2gf inversion:
  extended pairwise comparison between original ImageNet validation images and reconstructions obtained by minimizing Jeffreys divergence between
  pre-FC feature distributions plus a total-variation prior.}
  \label{fig:regnet-recon-ext}
\end{figure}

\end{document}